\documentclass{article}

     \usepackage[final]{neurips_2021}

\usepackage[utf8]{inputenc} 
\usepackage[T1]{fontenc}    
\usepackage{hyperref}       
\usepackage{url}            
\usepackage{booktabs}       
\usepackage{amsfonts}       
\usepackage{nicefrac}       
\usepackage{microtype}      
\usepackage{xcolor}         

\usepackage{amsthm}

\usepackage{threeparttable}

\usepackage{amsmath}
\usepackage{amssymb}

\usepackage{sidecap}

\usepackage{tabularx}
\usepackage{multirow}
\usepackage{hhline}
\usepackage{natbib}
\usepackage{algcompatible}
\usepackage{algorithm}
\usepackage{enumitem}
\usepackage{multicol}
\usepackage{bm}
\usepackage{svg}
\usepackage{titlesec}
\titlespacing\section{0pt}{0pt plus 0pt minus 1pt}{0pt plus 0pt minus 1pt}
\titlespacing\subsection{0pt}{0pt plus 0pt minus 1pt}{0pt plus 0pt minus 1pt}
\titlespacing\subsubsection{0pt}{0pt plus 0pt minus 1pt}{0pt plus 0pt minus 1pt}

\def\BW#1{\textbf{\textcolor{red}{#1}}}

\newcommand{\norm}[1]{\left\lVert#1\right\rVert}

\newtheorem{remark}{Remark}

\newtheorem{proposition}{Proposition}

\def\BW#1{{\textcolor{red}{#1}}}

\def\HJ#1{{\textcolor{orange}{#1}}}

\def\va{{\bm{a}}}
\def\vb{{\bm{b}}}

\def\ve{{\bm{e}}}

\def\vg{{\bm{g}}}
\def\vh{{\bm{h}}}

\def\vm{{\bm{m}}}

\def\vr{{\bm{r}}}
\def\vs{{\bm{s}}}

\def\vv{{\bm{v}}}

\def\vx{{\bm{x}}}
\def\vy{{\bm{y}}}
\def\vz{{\bm{z}}}

\def\mA{{\bm{A}}}

\def\mD{{\bm{D}}}

\def\mF{{\bm{F}}}

\def\mH{{\bm{H}}}
\def\mI{{\bm{I}}}
\def\mJ{{\bm{J}}}

\def\mL{{\bm{L}}}
\def\mM{{\bm{M}}}

\def\mP{{\bm{P}}}
\def\mQ{{\bm{Q}}}

\def\mU{{\bm{U}}}

\def\mW{{\bm{W}}}

\def \RR {{\mathbb{R}}}

\def \Mb {{\mathbf{M}}}

\usepackage{verbatim}

\usepackage{wrapfig}

\title{Heavy Ball Neural Ordinary Differential Equations}

\author{%
  Hedi Xia$^*$ \\
  Department of Mathematics\\ 
  University of California, Los Angeles\\
  \And
   Vai Suliafu \thanks{Co-first author} \\
   Scientific Computing and Imaging (SCI) Institute\\
   University of Utah, Salt Lake City, UT, USA\\
   \And
   Hangjie Ji \\
   Department of Mathematics\\ 
  University of California, Los Angeles\\
   \And
   Tan M. Nguyen\\
  Department of Mathematics\\ 
  University of California, Los Angeles\\
   \AND
   Andrea L. Bertozzi \\
  Department of Mathematics\\
  University of California, Los Angeles\\
   \And
   Stanley J. Osher \\
   Department of Mathematics\\
   University of California, Los Angeles\\
   \And
   Bao Wang \thanks{Please correspond to: wangbaonj@gmail.com} \\
   Department of Mathematics\\
   Scientific Computing and Imaging (SCI) Institute\\
   University of Utah, Salt Lake City, UT, USA\\
}

\begin{document}

\maketitle

\begin{abstract}
We propose heavy ball neural ordinary differential equations (HBNODEs), leveraging the continuous limit of the classical momentum accelerated gradient descent, to improve neural ODEs (NODEs) training and inference. HBNODEs have two properties that imply practical advantages over NODEs: (i) The adjoint state of an HBNODE also satisfies an HBNODE, accelerating both forward and backward ODE solvers, thus significantly reducing the number of function evaluations (NFEs) and improving the utility of the trained models. (ii) The spectrum of HBNODEs is well structured, enabling effective learning of long-term dependencies from complex sequential data. We verify the advantages of HBNODEs over NODEs on benchmark tasks, including image classification, learning complex dynamics, and sequential modeling. Our method requires remarkably fewer forward and backward NFEs, is more accurate, and learns long-term dependencies more effectively than the other ODE-based neural network models. Code is available at \url{https://github.com/hedixia/HeavyBallNODE}.

\end{abstract}

\section{Introduction}
Neural ordinary differential equations (NODEs) are a family of continuous-depth machine learning (ML) models whose forward and backward propagations rely on solving an ODE and its adjoint equation \citep{chen2018neural}. 
NODEs model the dynamics of hidden features $\vh(t)\in \RR^N$ using an ODE, 
which is parametrized by a neural network
$f(\vh(t),t,\theta)\in \RR^N$ with learnable parameters $\theta$, i.e.,
\begin{equation}\label{eq:NODE}
\frac{d\vh(t)}{dt} = f(\vh(t),t,\theta).
\end{equation}
\begin{wrapfigure}{r}{0.46\textwidth}
\vspace{-0.1cm}
\begin{center}
\begin{tabular}{cc}
\hskip -0.4cm\includegraphics[width=0.45\linewidth]{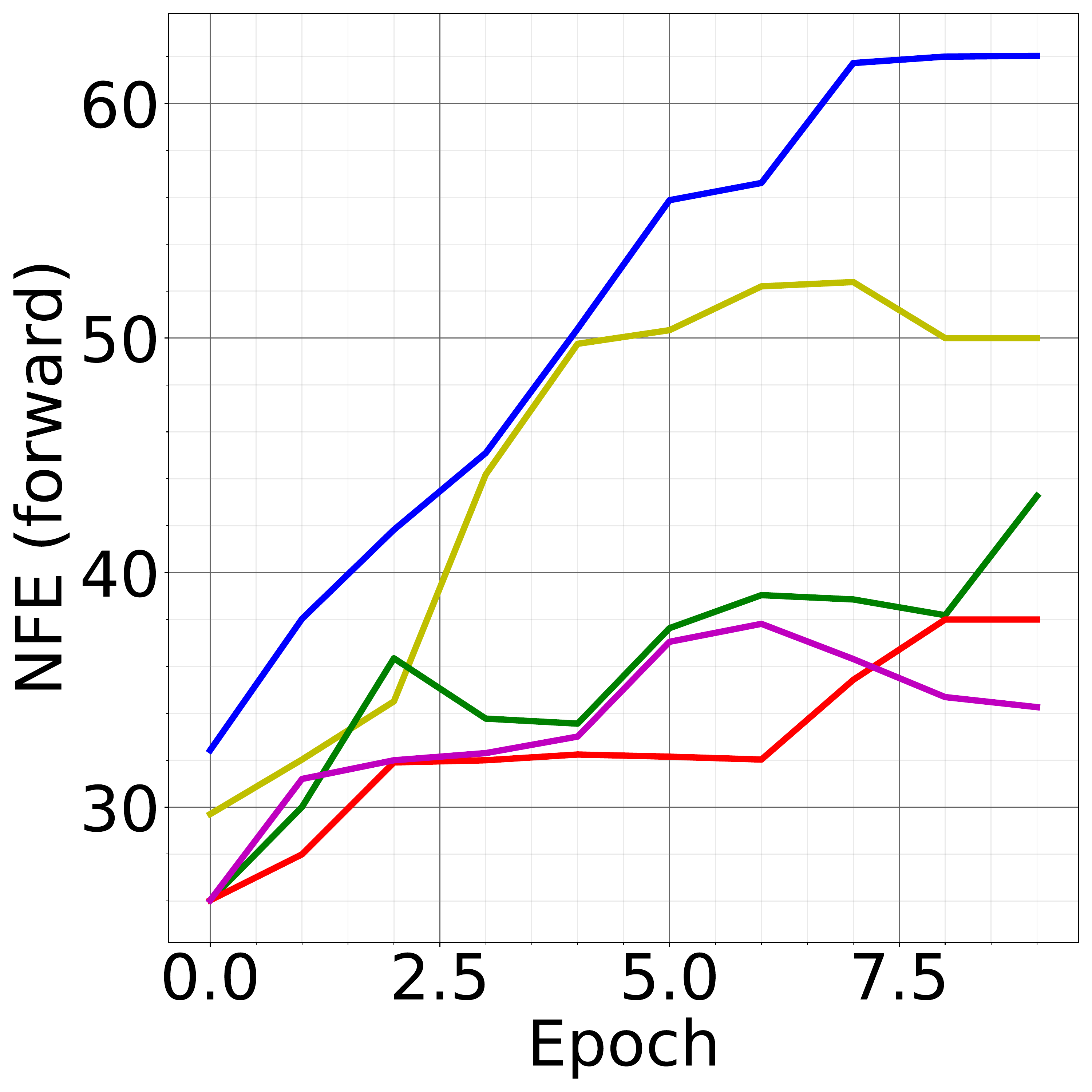}&
\hskip -0.4cm\includegraphics[width=0.45\linewidth]{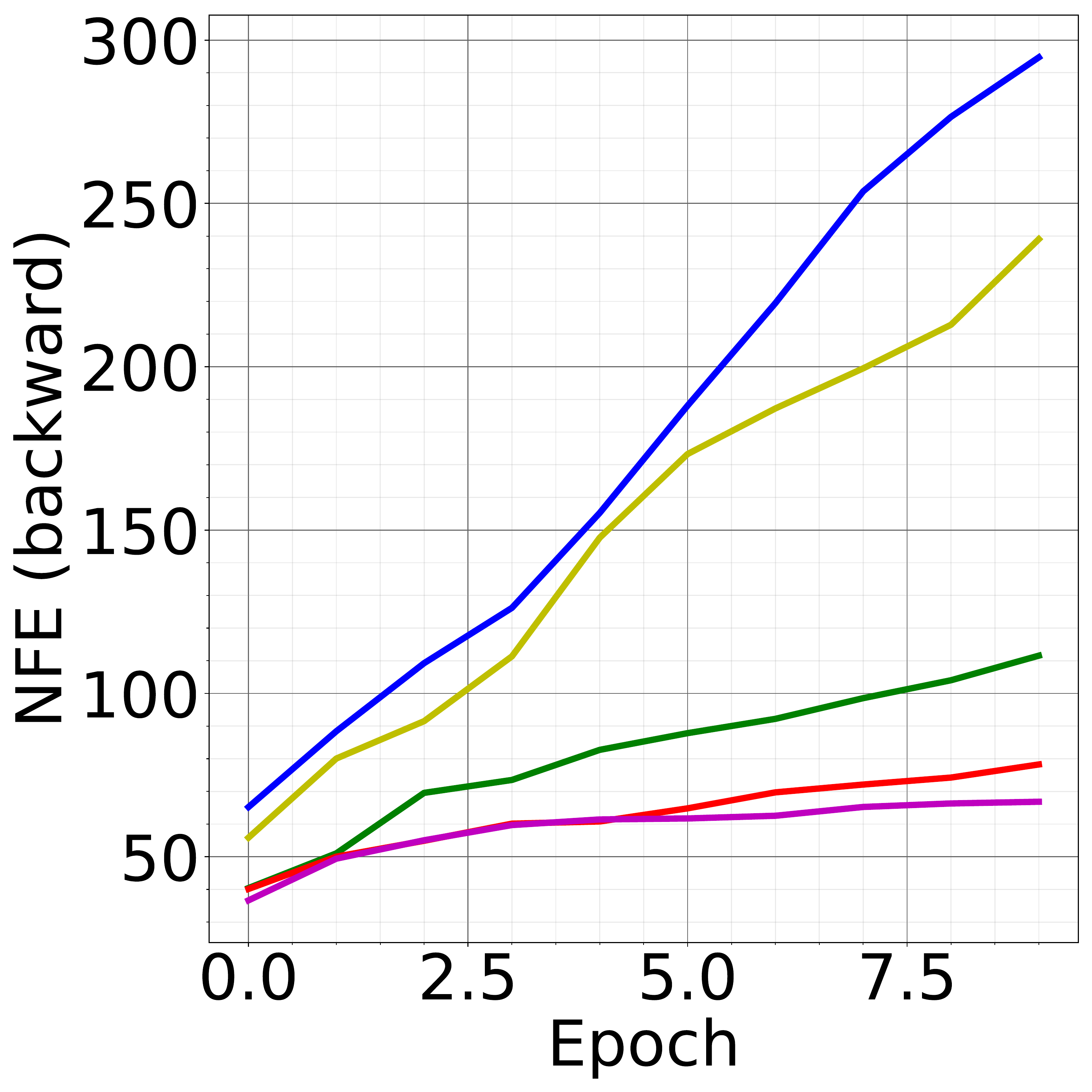}\\
\hskip -0.4cm\includegraphics[width=0.45\linewidth]{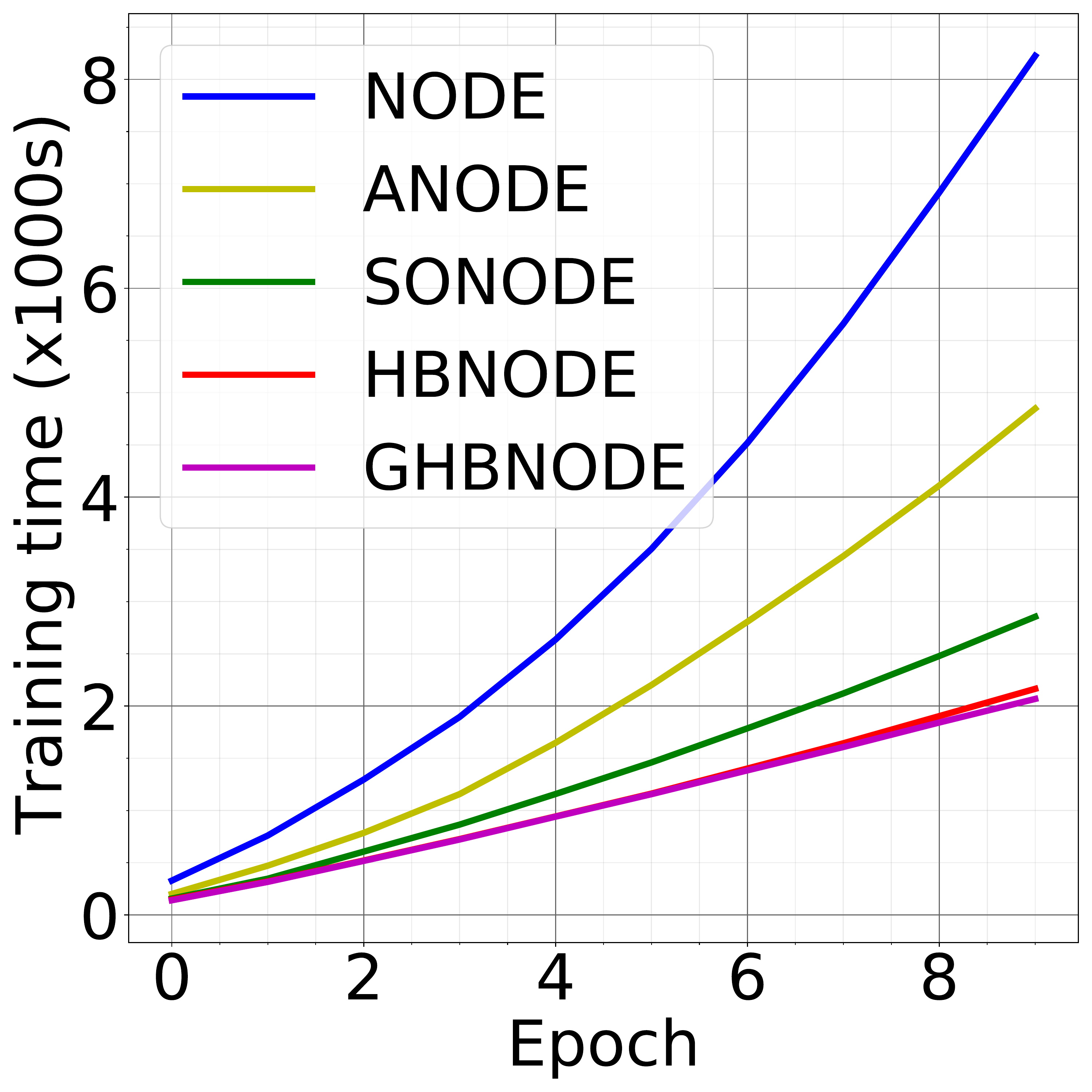}&
\hskip -0.4cm\includegraphics[width=0.45\linewidth]{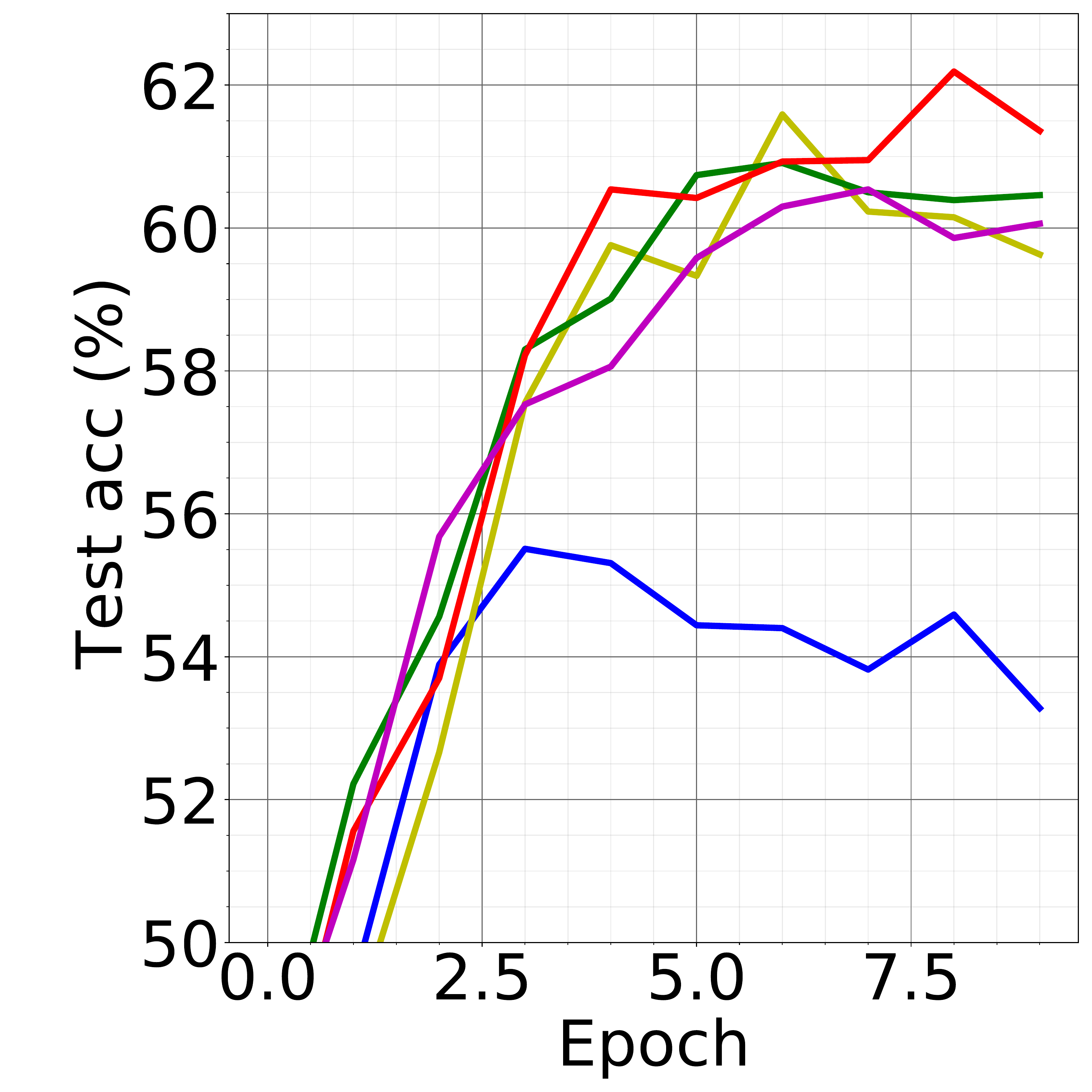}\\
  \end{tabular}
  \end{center}
  \vskip -0.2in
  \caption{Contrasting NODE, ANODE, SONODE, HBNODE, and GHBNODE for CIFAR10 classification in NFEs, training time, and test accuracy. (Tolerance: {$10^{-5}$}, see Sec.~\ref{sec:image-classification} for experimental details.)}\label{fig:node-hbnode-cifar10}
\end{wrapfigure}
Starting from the input $\vh(t_0)$, NODEs obtain the output $\vh(T)$ by solving \eqref{eq:NODE} for $t_0 \le t\le T$ with the initial value $\vh(t_0)$, using a black-box numerical ODE solver. 
The number of function evaluations (NFEs) that the black-box ODE solver
requires in a single forward pass is an analogue for the continuous-depth models \citep{chen2018neural} to the depth of networks in ResNets. 
The loss between NODE prediction $\vh(T)$ and the ground truth is denoted by $\mathcal{L}(\vh(T))$; we
update parameters $\theta$ using the following gradient
\citep{pontryagin2018mathematical}
\begin{equation}\label{eq:dL-dtheta}
\frac{d\mathcal{L}(\vh(T))}{d\theta} = \int_{t_0}^T \va(t)\frac{\partial f(\vh(t),t,\theta)}{\partial\theta}dt,
\end{equation}
where $\va(t):=\partial\mathcal{L}/\partial\vh(t)$ is the adjoint state, 
which satisfies the 
adjoint equation
\begin{equation}\label{eq:adjoint}
\frac{d\va(t)}{dt} = -\va(t)\frac{\partial f(\vh(t),t,\theta)}{\partial\vh}.
\end{equation}
NODEs are flexible in learning from
irregularly-sampled
sequential data and particularly suitable for learning 
complex 
dynamical systems \citep{chen2018neural,NEURIPS2019_42a6845a,NEURIPS2019_227f6afd,norcliffe2020_sonode,NEURIPS2020_e562cd9c,KidgerMFL20}, which can be trained by efficient algorithms \citep{
Quaglino2020SNODE:,NEURIPS2020_c24c6525,zhuang2021mali}. 
NODE-based continuous generative models have computational advantages over the 
classical normalizing flows \citep{chen2018neural,grathwohl2018scalable,NEURIPS2019_99a40143,pmlr-v119-finlay20a}. 
NODEs have also been generalized to neural stochastic differential 
equations, stochastic processes,
 and graph NODEs \citep{NEURIPS2019_59b1deff,pmlr-v108-li20i,poli2019graph,tzen2019neural,LG-ODE,norcliffe2021neural}. 
The drawback of NODEs is also prominent. In many ML tasks, NODEs require very high NFEs in both training and inference, especially in high accuracy settings where a lower tolerance is 
needed.
The NFEs increase rapidly with training; 
 high NFEs reduce computational speed and accuracy of NODEs 
and can lead to blow-ups in the worst-case scenario 
\citep{grathwohl2018scalable,NEURIPS2019_21be9a4b,massaroli2020dissecting,norcliffe2020_sonode}. As an illustration, we train NODEs for CIFAR10 classification using the same model and experimental settings as in \citep{NEURIPS2019_21be9a4b}, except using a tolerance of {$10^{-5}$}
; Fig.~\ref{fig:node-hbnode-cifar10} shows 
both forward and backward NFEs and the training time of different ODE-based models; we see that NFEs and computational
times increase
very rapidly for NODE, ANODE \citep{NEURIPS2019_21be9a4b}, and SONODE \citep{norcliffe2020_sonode}. More results on the large NFE and degrading utility issues for different benchmark experiments are available in Sec.~\ref{sec:experiments}. 
Another issue is that NODEs often fail to effectively learn long-term dependencies in sequential data \citep{lechner2020learning}, 
 discussed in Sec.~\ref{sec:lowerbounds}.


\subsection{Contribution}
We propose heavy ball neural ODEs (HBNODEs), 
leveraging the continuous limit of the classical momentum accelerated gradient descent, to improve NODE training and inference. At the core of HBNODE is replacing the first-order ODE \eqref{eq:NODE} with a heavy ball ODE (HBODE), i.e., a second-order ODE with an
appropriate damping term. HBNODEs 
have two theoretical
properties that imply 
practical advantages over NODEs: 
\begin{itemize}[leftmargin=*]
    \item The adjoint equation 
    used for training a HBNODE is also a HBNODE
    (see Prop.~\ref{prop:adjoint-HBNODE} {and Prop.~}\ref{prop:adjoint-HBNODE-1st}), accelerating both forward and backward propagation, thus significantly 
    reducing
    both forward and backward NFEs.
    The reduction in NFE using HBNODE over existing benchmark ODE-based  models becomes more aggressive as the error tolerance of the ODE solvers decreases.
    \item The spectrum of the HBODE is well-structured (see Prop.~\ref{lemma-eigan-M}),
     alleviating 
    the vanishing gradient issue in back-propagation and 
    {enabling the model to effectively learn}
    long-term dependencies from sequential data. 
\end{itemize}
To mitigate the potential 
{blow-up problem}
in training 
{HBNODEs},
we further propose generalized HBNODEs (GHBNODEs) by integrating 
skip connections \citep{he2016identity} and gating mechanisms \citep{hochreiter1997long}
into the HBNODE. See Sec.~\ref{sec:GHBNODEs} for details. 

\subsection{Organization}
We organize the paper as follows: In Secs~\ref{sec:HBNODE} and ~\ref{sec:GHBNODEs}, we present our motivation, algorithm, and analysis of HBNODEs and GHBNODEs, respectively. 
We analyze the spectrum structure of the adjoint equation of HBNODEs/GHBNODEs in Sec.~\ref{sec:lowerbounds}, 
which indicates
that HBNODEs/GHBNODEs can learn long-term dependency effectively. 
We test the performance of HBNODEs and GHBNODEs on benchmark point cloud separation, image classification, learning dynamics, 
and sequential modeling in Sec.~\ref{sec:experiments}. 
We discuss more related work in Sec.~\ref{sec:related-works}, followed by concluding remarks.
Technical proofs and 
more experimental details are provided in the appendix.

\section{Heavy Ball Neural Ordinary Differential Equations}\label{sec:HBNODE}
\subsection{Heavy ball ordinary differential equation}
Classical momentum method, a.k.a., the heavy ball method, has achieved remarkable success in accelerating gradient descent \citep{polyak1964some} and 
has significantly improved the training of deep neural networks
\citep{sutskever2013importance}. 
As the continuous limit of the classical momentum method, heavy ball ODE (HBODE) has been studied in various settings and 
{has been used to analyze}
the acceleration phenomenon of the momentum methods.
For the ease of reading and completeness, we derive the HBODE from the 
classical momentum method. 
Starting from initial points $\vx^0$ and $\vx^1$, gradient descent with classical momentum searches a minimum of the function $F(\vx)$ through the following iteration
\begin{equation}\label{eq:Appendix-HB1}
\vx^{k+1} = \vx^k - s \nabla F(\vx^k) + \beta (\vx^k-\vx^{k-1}),
\end{equation}
where $s>0$ is the step size and $0\leq \beta <1$ is the momentum hyperparameter. 
For any fixed step size $s$, let 
$
\vm^k := ({\vx^{k+1}-\vx^k})/{\sqrt{s}},
$ 
{and
let $\beta := 1-\gamma \sqrt{s}$,
where $\gamma\geq 0$ is another hyperparameter.}
{Then} we can rewrite \eqref{eq:Appendix-HB1} as 
\begin{equation}\label{eq:Appendix-HB3}
\vm^{k+1}=(1-\gamma\sqrt{s})\vm^k -\sqrt{s}\nabla F(\vx^k);\ \vx^{k+1} = \vx^k + \sqrt{s}\vm^{k+1}.
\end{equation}
 Let 
$s\to 0$ in \eqref{eq:Appendix-HB3}; we obtain the following system of 
first-order ODEs, 
\begin{equation}\label{eq:Appendix-HBNODE1}
\frac{d\vx(t)}{dt}= \vm(t);\ \frac{d\vm(t)}{dt} = -\gamma \vm(t) - \nabla F(\vx(t)). 
\end{equation}
This can be further 
{rewritten}
as a
second-order {heavy ball ODE (HBODE), which also models a damped oscillator,}
\begin{equation}\label{eq:Appendix-HBNODE2}
\frac{d^2\vx(t)}{dt^2} + \gamma \frac{d\vx(t)}{dt} = -\nabla F(\vx(t)).
\end{equation}

\begin{wrapfigure}{r}{0.46\textwidth}
\vspace{-0.6cm}
\begin{center}
\begin{tabular}{c}
\includegraphics[width=\linewidth]{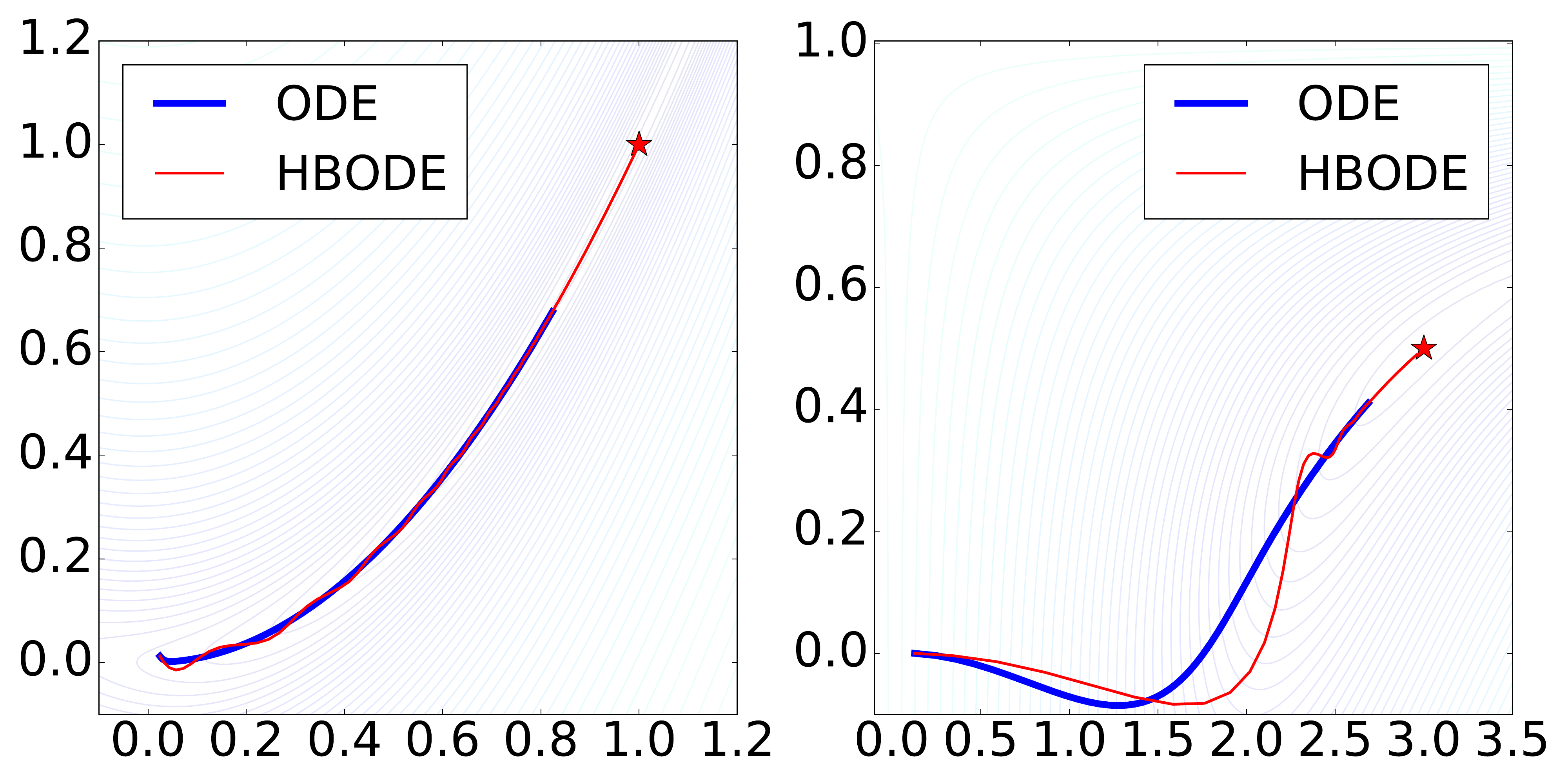}\\[-3pt]
  \end{tabular}
  \end{center}
  \vskip -0.15in
\caption{ Comparing the trajectory of ODE 
and HBODE 
when $F(\vx)$ is the Rosenbrock (left) and Beale (right) functions.  
}\label{fig:compare-dynamics-benchmarks}\vspace{-0.15in}
\end{wrapfigure}
We compare the dynamics of HBODE \eqref{eq:Appendix-HBNODE2} and the following ODE 
limit of the gradient descent (GD)
\begin{equation}\label{eq:ODE-HBODE}
\frac{d\vx}{dt} = -\nabla F(\vx).
\end{equation}

In particular, we 
solve the ODEs \eqref{eq:Appendix-HBNODE2} and \eqref{eq:ODE-HBODE} with $F(\vx)$ defined as a \texttt{Rosenbrock} \citep{10.1093/comjnl/3.3.175} or \texttt{Beale} \citep{10.1145/800184.810512} function (see Appendix~\ref{appendix-compare-dynamics-odes} for experimental details). Fig.~\ref{fig:compare-dynamics-benchmarks} shows that 
{with} the same 
numerical ODE solver, 
HBODE 
converges to the stationary point
{(marked by stars)}
faster than \eqref{eq:ODE-HBODE}. 
The fact that HBODE can 
accelerate the dynamics of the ODE for a gradient system motivates us to propose HBNODE to accelerate forward propagation of NODE.


\subsection{Heavy ball neural ordinary differential equations}
{Similar to NODE, we parameterize $-\nabla F$ in \eqref{eq:Appendix-HBNODE2} using a neural network $f(\vh(t),t,\theta)$,} resulting in the following HBNODE
with initial position $\vh(t_0)$ and 
momentum $\vm(t_0):=d\vh/dt(t_0)$, 
\begin{equation}\label{eq:HBNODE-2nd}
\frac{d^2\vh(t)}{dt^2} + \gamma \frac{d\vh(t)}{dt} = f(\vh(t),t,\theta), 
\end{equation}
where $\gamma\geq 0$ is the damping parameter, which can be set as a tunable or a learnable hyperparmater with positivity constraint. 
In {the} trainable case, we use $\gamma = \epsilon \cdot \text{sigmoid}(\omega)$ for a trainable $\omega\in \mathbb{R}$ and a fixed tunable upper bound $\epsilon$ (we set $\epsilon=1$ below).
According to \eqref{eq:Appendix-HBNODE1}, HBNODE \eqref{eq:HBNODE-2nd} is equivalent to 
\begin{equation}\label{eq:HBNODE-1st}
\frac{d\vh(t)}{dt} = \vm(t); 
\quad \frac{d\vm(t)}{dt} = -\gamma \vm(t) + f(\vh(t),t,\theta).
\end{equation}
Equation \eqref{eq:HBNODE-2nd}
{(or equivalently, the system \eqref{eq:HBNODE-1st})}
defines the forward ODE 
for the HBNODE, {and} we can use either the first-order (Prop.~\ref{prop:adjoint-HBNODE-1st}) or the second-order (Prop.~\ref{prop:adjoint-HBNODE}) adjoint sensitivity method to update the parameter $\theta$ \citep{norcliffe2020_sonode}. 

\begin{proposition}[Adjoint equation for HBNODE]\label{prop:adjoint-HBNODE}
The adjoint state $\va(t):=\partial\mathcal{L}/\partial\vh(t)$ for the HBNODE \eqref{eq:HBNODE-2nd} satisfies the following HBODE with the same damping parameter $\gamma$ as that in \eqref{eq:HBNODE-2nd}, 
\begin{equation}\label{eq:adjoint-HBNODE-2nd}
\frac{d^2\va(t)}{dt^2} - \gamma\frac{d\va(t)}{dt} = \va(t) \frac{\partial f}{\partial\vh}(\vh(t),t,\theta). 
\end{equation}
\end{proposition}
\begin{remark}\label{remark-adjoint-HBNODE-2nd}
Note that we solve the adjoint equation \eqref{eq:adjoint-HBNODE-2nd} from time 
{$t=T$ to $t=t_0$}
in the backward propagation{. By}
letting $\tau=T-t$ and $\vb(\tau)=\va(T-\tau)$, we 
can 
rewrite \eqref{eq:adjoint-HBNODE-2nd} as follows,
\begin{equation}\label{eq:adjoint-HBNODE-2nd-b}
\frac{d^2\vb(\tau)}{d\tau^2} + \gamma \frac{d\vb(\tau)}{d\tau} = \vb(\tau)\frac{\partial f}{\partial\vh}(\vh(T-\tau),T-\tau,\theta).
\end{equation}
Therefore, the adjoint of the HBNODE is also a HBNODE and they have the same damping parameter.
\end{remark}
We can also employ \eqref{eq:HBNODE-1st} and its adjoint for the forward and backward propagations, respectively. 
\begin{proposition}[Adjoint equations for the first-order HBNODE system]\label{prop:adjoint-HBNODE-1st}
The adjoint states $\va_\vh(t)$ $:=\partial\mathcal{L}/\partial\vh(t)$ and  $\va_\vm(t):=\partial\mathcal{L}/\partial\vm(t)$ for the first-order HBNODE system \eqref{eq:HBNODE-1st} 
{satisfy}
\begin{equation}\label{eq:adjoint-HBNODE-1st}
\frac{d\va_\vh(t)}{dt} = -\va_\vm(t) \frac{\partial f}{\partial\vh}(\vh(t),t,\theta); \quad \frac{d\va_\vm(t)}{dt} = -\va_\vh(t) + \gamma \va_\vm(t).
\end{equation}
\end{proposition}
\begin{remark}\label{remark-adjoint-HBNODE-1st}
Let $\tilde{\va}_\vm(t)=d\va_\vm(t)/dt$, then $\va_\vm(t)$ and $\tilde{\va}_\vm(t)$ satisfies the following first-order heavy ball ODE system 
\begin{equation}\label{eq:Adjoint-firsrt-order-system}
\frac{d\va_\vm(t)}{dt} =  \tilde{\va}_\vm(t); \quad \frac{d\tilde{\va}_\vm(t)}{dt} = \va_\vm(t)\frac{\partial f}{\partial \vh}(\vh(t),t,\theta) + \gamma \tilde{\va}_\vm(t).
\end{equation}
{Note that we solve this system backward in time in back-propagation.}
Moreover, 
we have $\va_{\vh}(t) = \gamma \va_\vm(t) - \tilde{\va}_\vm(t)$.
\end{remark}
Similar to \citep{norcliffe2020_sonode}, we use the coupled first-order HBNODE system \eqref{eq:HBNODE-1st} and its adjoint first-order HBNODE system \eqref{eq:adjoint-HBNODE-1st} for practical implementation, since the entangled representation permits 
faster computation \citep{norcliffe2020_sonode} of the gradients 
{of}
the coupled ODE systems. 



\section{Generalized Heavy Ball Neural Ordinary Differential Equations}\label{sec:GHBNODEs}

\begin{wrapfigure}{r}{.45\textwidth}
\includegraphics[width=\linewidth]{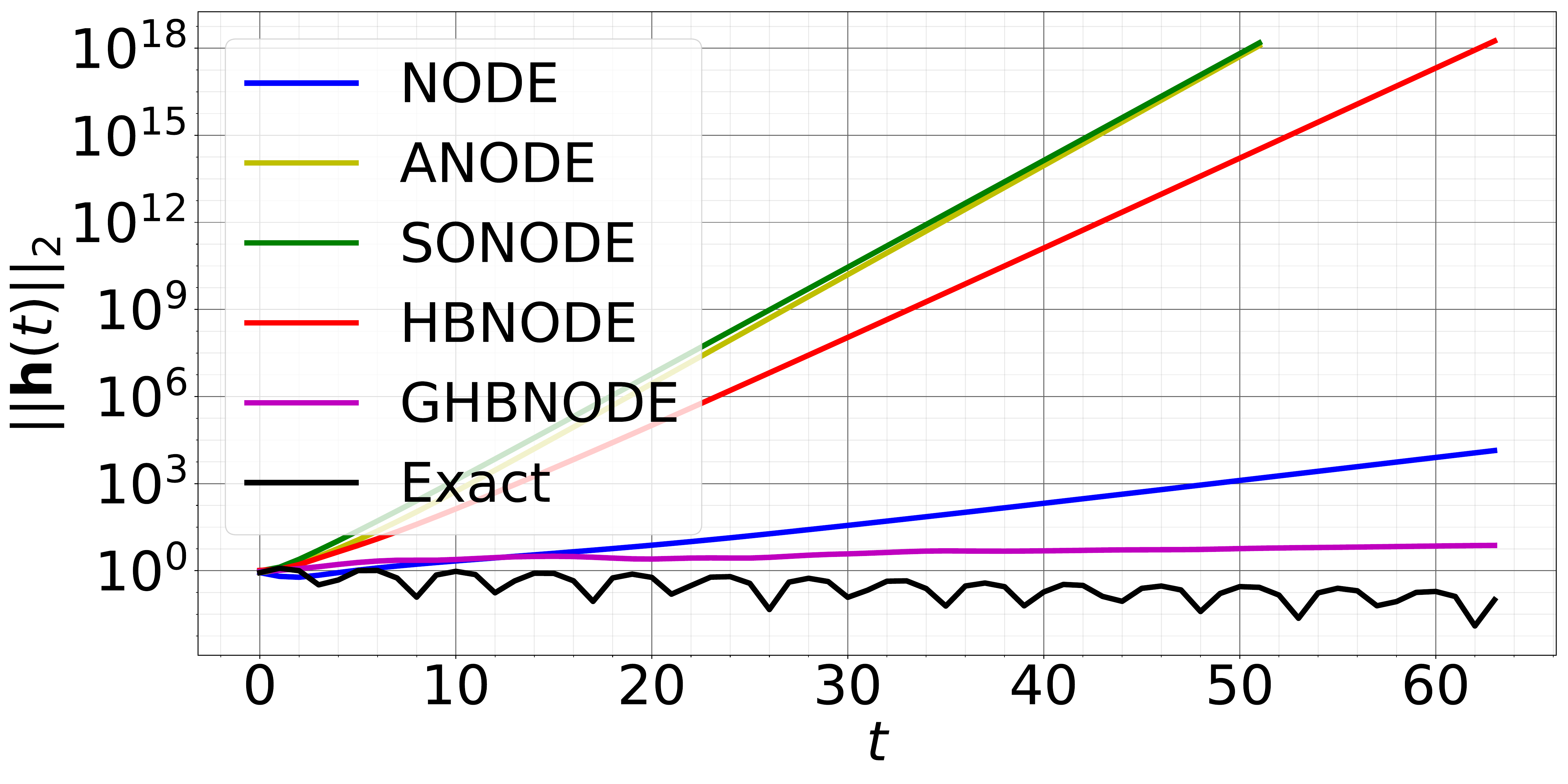}
\vspace{-0.2in}
\caption{ Contrasting 
$\vh(t)$ for different models. $\vh(t)$ in ANODE, SONODE, and HBNODE grows much faster than that in NODE. GHBNODE controls the growth of $\vh(t)$ effectively 
when $t$ is large.}\label{fig:blow-up}
\end{wrapfigure}

In this section, we propose a generalized version of HBNODE (GHBNODE), see \eqref{eq:GHBNODE}, to mitigate the potential blow-up issue in training ODE-based models. 
In our experiments, we {observe}
that $\vh(t)$ of ANODEs \citep{NEURIPS2019_21be9a4b}, SONODEs \citep{norcliffe2020_sonode}, and HBNODEs \eqref{eq:HBNODE-1st} usually grows much faster than that of NODEs. The fast growth of $\vh(t)$ can lead to finite-time blow up.  
As an illustration, 
we compare 
{the performance of}
NODE, ANODE, SONODE, HBNODE, and GHBNODE on {the} Silverbox task as in \citep{norcliffe2020_sonode}. 
{The goal of the task is to}
learn the voltage of an electronic circuit 
{that resembles}
a Duffing oscillator, 
{where the}
input voltage $V_1(t)$ 
{is used}
to predict 
{the}
output $V_2(t)$. Similar to
{the setting in} \citep{norcliffe2020_sonode}, we first augment ANODE by 1 dimension with 0-augmentation and augment SONODE, HBNODE, and GHBNODE with a dense network.
We use a simple dense layer to parameterize $f$ for all five models,  
with an extra input term for $V_1(t)$\footnote{Here, we exclude an $\vh^3$ term that appeared in the original Duffing oscillator model
because including it would result in finite-time explosion.}.
For both HBNODE and GHBNODE, we set the damping parameter $\gamma$ to be 
${\rm sigmoid}(-3) 
$. For GHBNODE \eqref{eq:GHBNODE} below, 
we set $\sigma(\cdot)$ to be the \texttt{hardtanh} function with bound $[-5,5]$ and $\xi=\ln(2)$. 
The detailed architecture can be found in Appendix~\ref{appendix:experimental-details}.  
As shown in Fig.~\ref{fig:blow-up},
{compared to the vanilla NODE, the $\ell_2$ norm of $\vh(t)$ grows much faster when a higher order NODE is used, which leads to blow-up during training.}
{Similar issues arise in the time series experiments (see Sec.~\ref{sec:sequential-modeling}), where SONODE blows up during long term integration in time, and HBNODE suffers from the same issue with some initialization.}

To alleviate {the problem above,}
we propose the following generalized HBNODE 
\begin{equation}\label{eq:GHBNODE}
\begin{aligned}
\frac{d\vh(t)}{d t} &= \sigma(\vm(t)),\\
\frac{d\vm(t)}{d t} &= -\gamma\vm(t) + f(\vh(t),t,\theta) - \xi\vh(t), 
\end{aligned}
\end{equation}
where $\sigma(\cdot)$ is a nonlinear activation,
which is set as $\tanh$ in our experiments.
{The positive hyperparameters}
$\gamma,\xi>0$ are 
tunable or learnable.
In {the}
trainable case, we let $\gamma = \epsilon \cdot \text{sigmoid}(\omega)$ as in HBNODE, and $\xi = \text{softplus}(\chi)$ to ensure that $\gamma,\xi \geq 0$. Here, we integrate two main ideas into the {design of} GHBNODE: 
(i) {We incorporate} the gating mechanism 
used in LSTM \citep{hochreiter1997long} and GRU \citep{cho2014learning}, which can suppress the  
aggregation of $\vm(t)$;
(ii) 
{Following}
the idea of skip connection \citep{he2016identity}, we add the term $\xi\vh(t)$ into the governing equation of $\vm(t)$, which benefits training and generalization of GHBNODEs. Fig.~\ref{fig:blow-up} shows that GHBNODE can indeed control the growth of $\vh(t)$ effectively. 
\begin{proposition}[Adjoint equations for GHBNODEs]\label{prop:adjoint-GHBNODE}
The adjoint states $\va_\vh(t):=\partial\mathcal{L}/\partial\vh(t)$, $\va_\vm(t):=\partial\mathcal{L}/\partial\vm(t)$ for the GHBNODE \eqref{eq:GHBNODE} 
{satisfy}
the following first-order ODE system
\begin{equation}\label{eq:adjoint-GHBNODE-2nd}
\frac{\partial \va_\vh(t)}{\partial t} = - \va_\vm(t)\Big(\frac{\partial f}{\partial \vh}(\vh(t),t,\theta) - \xi\mI\Big),
\quad 
\frac{\partial \va_\vm(t)}{\partial t} = -\va_\vh(t) \sigma'(\vm(t)) + \gamma \va_\vm(t).
\end{equation}
\end{proposition}
Though the adjoint state of the GHBNODE \eqref{eq:adjoint-GHBNODE-2nd} does not satisfy the exact  heavy ball ODE, based on our empirical study, it also significantly reduces the backward NFEs.

\section{Learning long-term dependencies -- Vanishing 
gradient 
}\label{sec:lowerbounds}
It is known that the vanishing and exploding gradients are two bottlenecks for training recurrent neural networks (RNNs) with long-term dependencies \citep{bengio1994learning,pascanu2013difficulty} (see Appendix~\ref{appendix-vanishing-gradent} for a brief review on the exploding and vanishing gradient issues in training RNNs). The exploding gradients issue can be effectively resolved via gradient clipping, training loss regularization, etc \citep{pascanu2013difficulty,erichson2021lipschitz}. Thus in practice the vanishing gradient is the major issue for learning long-term dependencies \citep{pascanu2013difficulty}. As the continuous analogue of RNN,
NODEs as well as their hybrid ODE-RNN models, may {also} suffer from vanishing in the adjoint state $\va(t):=\partial\mathcal{L}/\partial\vh(t)$ \citep{lechner2020learning}. When {the} vanishing gradient issue happens, $\va(t)$ goes to ${\bf 0}$ quickly as $T-t$ increases, then $d\mathcal{L}/d\theta$ in \eqref{eq:dL-dtheta} will be 
independent of these $\va(t)$. We have the following expressions for the adjoint states of the NODE and HBNODE {(see Appendix~\ref{appendix-vanishing-gradent} for detailed derivation)}:
\begin{itemize}[leftmargin=*]
\item For NODE, we have 
\begin{equation}\label{eq:NODE-gradient}
\frac{\partial\mathcal{L}}{\partial\vh_t} = \frac{\partial\mathcal{L}}{\partial\vh_T}\frac{\partial\vh_T}{\partial\vh_t} = \frac{\partial\mathcal{L}}{\partial\vh_T}\exp\Big\{-\int_T^t\frac{\partial f}{\partial \vh}(\vh(s),s,\theta)ds\Big\}.
\end{equation}
\item For GHBNODE\footnote{HBNODE can be seen as a special GHBNODE with $\xi=0$ and $\sigma$ be the identity map.
}, 
from \eqref{eq:adjoint-HBNODE-1st} we can derive 
\begin{equation}\label{eq:HBNODE-gradient}
{\small \begin{bmatrix}
\frac{\partial\mathcal{L}}{\partial\vh_t}&\hspace{-0.1in} \frac{\partial\mathcal{L}}{\partial\vm_t} 
\end{bmatrix} = 
\begin{bmatrix}
\frac{\partial\mathcal{L}}{\partial\vh_T}&\hspace{-0.1in}  \frac{\partial\mathcal{L}}{\partial\vm_T} 
\end{bmatrix}
\begin{bmatrix}
\frac{\partial\vh_T}{\partial\vh_t} &\hspace{-0.1in} \frac{\partial\vh_T}{\partial\vm_t}\\
\frac{\partial\vm_T}{\partial\vh_t} &\hspace{-0.1in}
\frac{\partial\vm_T}{\partial\vm_t}\\
\end{bmatrix}=
\begin{bmatrix}
\frac{\partial\mathcal{L}}{\partial\vh_T} \  \frac{\partial\mathcal{L}}{\partial\vm_T} \end{bmatrix}\exp\Big\{-\underbrace{\int_T^t\begin{bmatrix}
{\bf 0} &\hspace{-0.05in} \frac{\partial \sigma}{\partial \vm}\\
\big(\frac{\partial f}{\partial\vh}-\xi\mI\big) &\hspace{-0.05in}
-\gamma\mI
\end{bmatrix}ds}_{:=\mM} \Big\}.}
\end{equation}
\end{itemize}

Note that 
the matrix exponential is directly related to its eigenvalues. 
{By}
Schur decomposition, there exists an orthogonal matrix $\mQ$ and {an} upper triangular matrix $\mU$, 
{where the diagonal entries of $\mU$ are eigenvalues of $\mQ$ ordered by their real parts,}
such that
\begin{equation}
    -\mM = \mQ\mU\mQ^\top \Longrightarrow 
    \exp\{-\mM\} = \mQ\exp\{\mU\}\mQ^\top. 
\end{equation}
Let $\vv^\top := \begin{bmatrix}\frac{\partial\mathcal{L}}{\partial\vh_T} \ \frac{\partial\mathcal{L}}{\partial\vm_T} \end{bmatrix}\mQ$, then \eqref{eq:HBNODE-gradient} can be rewritten as 
\begin{equation}\label{eq:sgfsfxvb}
\begin{bmatrix}
\frac{\partial\mathcal{L}}{\partial\vh_t} \  \frac{\partial\mathcal{L}}{\partial\vm_t} 
\end{bmatrix} =
\begin{bmatrix}
\frac{\partial\mathcal{L}}{\partial\vh_T} \  \frac{\partial\mathcal{L}}{\partial\vm_T} \end{bmatrix}\exp\{-\mM\} = 
\begin{bmatrix}
\frac{\partial\mathcal{L}}{\partial\vh_T} \  \frac{\partial\mathcal{L}}{\partial\vm_T} \end{bmatrix}\mQ\exp\{\mU\}\mQ^\top = 
\vv^\top \exp\{\mU\}\mQ^\top.
\end{equation}
By taking the $\ell_2$ norm in \eqref{eq:sgfsfxvb} and dividing both sides by $\norm{\begin{bmatrix}\frac{\partial\mathcal{L}}{\partial\vh_T} \ \frac{\partial\mathcal{L}}{\partial\vm_T} \end{bmatrix}}_2$, we arrive at
\begin{equation}\label{eq:ratio:HBNODE}
\frac{\norm{\begin{bmatrix}
\frac{\partial\mathcal{L}}{\partial\vh_t} \  \frac{\partial\mathcal{L}}{\partial\vm_t} 
\end{bmatrix}}_2}
{\norm{\begin{bmatrix}\frac{\partial\mathcal{L}}{\partial\vh_T} \ \frac{\partial\mathcal{L}}{\partial\vm_T} \end{bmatrix}}_2}
= 
\frac{\norm{\vv^\top \exp\{\mU\}\mQ^\top}_2}{\norm{\vv^\top\mQ^\top}_2}
= 
\frac{\norm{\vv^\top \exp\{\mU\}}_2}{\norm{\vv}_2}
=
\norm{\ve^\top \exp\{\mU\}}_2,
\end{equation}
i.e., $\norm{\begin{bmatrix}
\frac{\partial\mathcal{L}}{\partial\vh_t} \  \frac{\partial\mathcal{L}}{\partial\vm_t} 
\end{bmatrix}}_2=\norm{\ve^\top \exp\{\mU\}}_2 \norm{\begin{bmatrix}\frac{\partial\mathcal{L}}{\partial\vh_T} \ \frac{\partial\mathcal{L}}{\partial\vm_T} \end{bmatrix}}_2$ where $\ve = {\vv}/{\norm{\vv}_2}$. 

\begin{proposition}\label{lemma-eigan-M}
The eigenvalues of 
$-\mM$ 
can be paired so that the sum of each pair equals 
$(t-T)\gamma$. 
\end{proposition} 

 
For a given constant $a>0$, 
we can group the upper triangular matrix $\exp\{\mU\}$ as follows
\begin{equation}
\exp\{\mU\}:= \begin{bmatrix}\exp\{\mU_L\} & \mP 
\\ {\bf 0} & \exp\{\mU_V\}\end{bmatrix},
\end{equation}
where the diagonal of $\mU_L$ ($\mU_V$) contains eigenvalues of $-\mM$ that are no less (greater) than $(t-T)a$.
Then, we have $\|\ve^\top \exp\{\mU\}\|_2 \geq \|\ve_L^\top \exp\{\mU_L\}\|_2$ where the vector $\ve_L$ denotes the first $m$ columns
of $\ve$ with $m$ be the number of columns of $\mU_L$. By choosing {$0\leq \gamma \leq 2a$,} 
for every 
pair of eigenvalues of {$-\mM$} 
there is at least one eigenvalue whose real part is no less than $(t-T)a$. 
{Therefore, $\exp\{\mU_L\}$ decays at a rate at most $(t-T)a$, and the dimension of $\mU_L$ is at least $N\times N$.
}
We avoid exploding gradients by clipping the $\ell_2$ norm of the adjoint states similar to that used for training RNNs. 

\begin{figure}
\begin{center}
  \includegraphics[clip, trim=0.1cm 0.1cm 3.9cm 0.1cm,height=3.5cm]{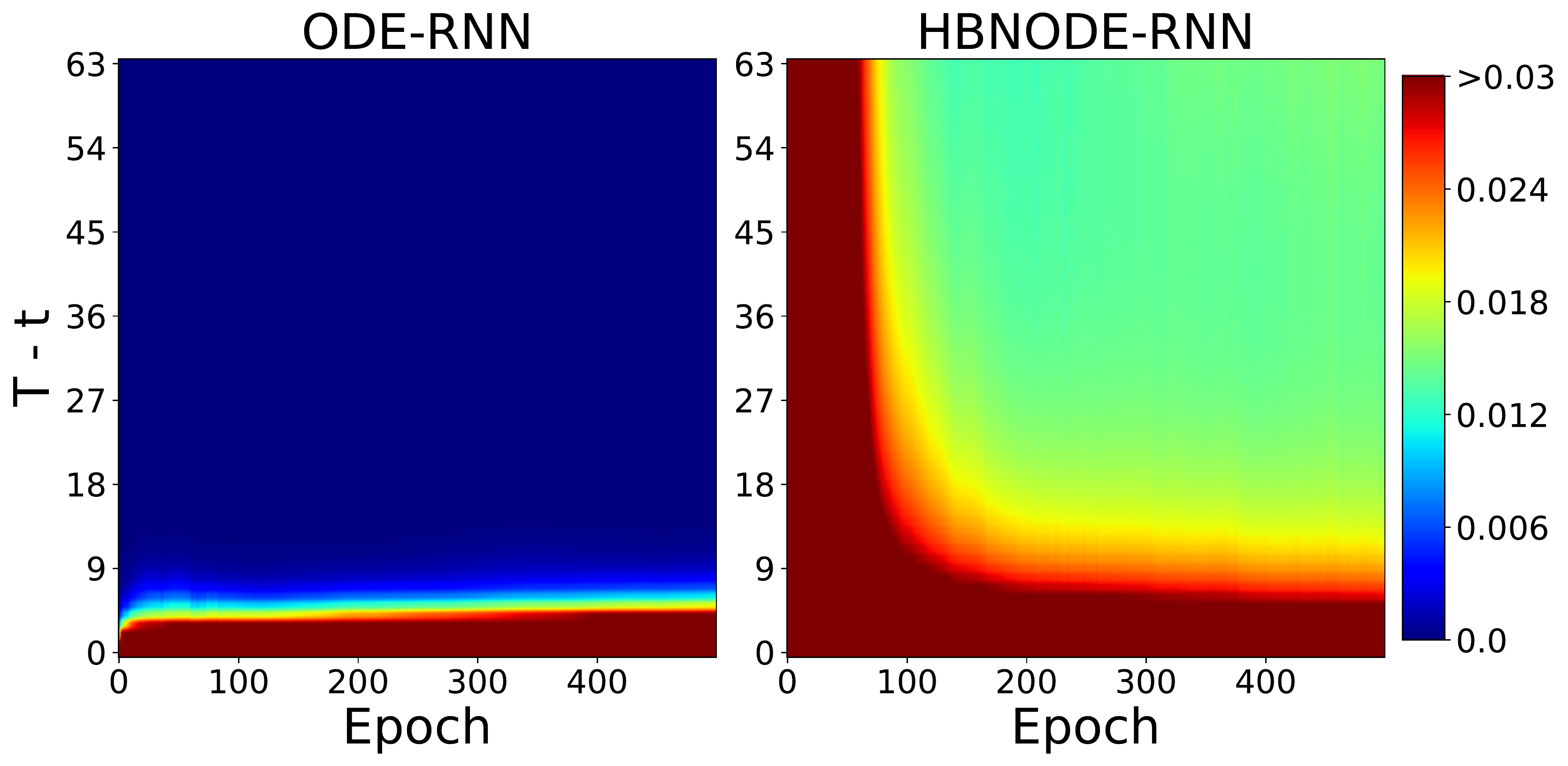}
  \includegraphics[clip, trim=16.5cm 0.1cm 0.1cm 0.1cm,height=3.5cm]{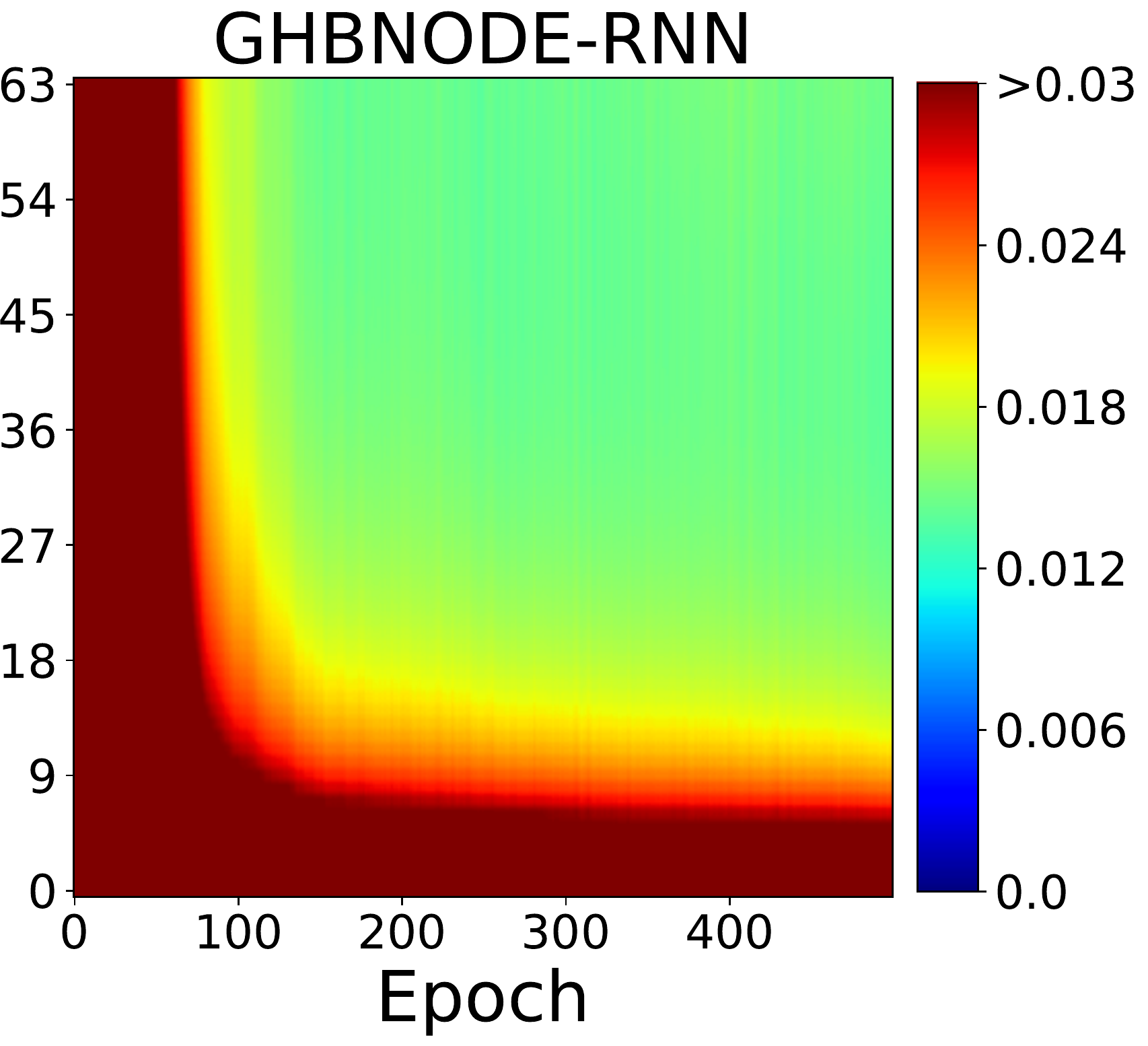}\label{fig:gradient-vanishing}
  \vspace{-3mm}
   \caption{Plot of the the $\ell_2$-norm of the adjoint states for ODE-RNN and (G)HBNODE-RNN back-propagated from the last time stamp. The adjoint state of ODE-RNN vanishes quickly when the gap between the final time $T$ and intermediate time $t$ becomes larger, while the adjoint states of (G)HBNODE-RNN decays much more slowly. This implies that (G)HBNODE-RNN is more effective in learning long-term dependency
   than
   ODE-RNN. 
  }
\end{center}
\end{figure}

In contrast, all eigenvalues of the matrix $\int_T^t {\partial f}/{\partial \vh}ds$ in \eqref{eq:NODE-gradient} {for NODE} can be very positive or negative, 
resulting in exploding or vanishing
{gradients}.
As an illustration, we consider the benchmark Walker2D kinematic simulation task that requires learning long-term dependencies effectively \citep{lechner2020learning,brockman2016openai}. We train ODE-RNN \citep{NEURIPS2019_42a6845a} and (G)HBNODE-RNN on this benchmark dataset, and the detailed experimental settings are provided in Sec.~\ref{sec:sequential-modeling}. Figure~\ref{fig:gradient-vanishing} plots $\|\partial\mathcal{L}/\partial\vh_{t}\|_2$ for ODE-RNN and $\|[\partial\mathcal{L}/\partial \vh_{t}\ \partial\mathcal{L}/\partial\vm_{t}] \|_2$
{for (G)HBNODE-RNN}, showing that the adjoint state of ODE-RNN vanishes quickly, while {that of} (G)HBNODE-RNN 
{does not}
vanish even when the gap between $T$ and $t$ is very large.



\section{Experimental Results}\label{sec:experiments}
In this section, we compare the performance of the proposed HBNODE and GHBNODE with existing ODE-based models, including NODE \citep{chen2018neural}, ANODE \citep{NEURIPS2019_21be9a4b}, and SONODE \citep{norcliffe2020_sonode} on the benchmark 
point cloud separation, 
image classification, learning dynamical systems, and kinematic simulation. 
For all the experiments, we use Adam \citep{kingma2014adam} as the benchmark optimization solver (the learning rate and batch size for each experiment 
{are}
listed in Table~\ref{Tab:batch-size-learning-rate}) and Dormand–Prince-45 as the numerical ODE solver. 
For HBNODE and GHBNODE, we 
{set}
{$\gamma={\rm sigmoid}(\theta)$, }
{where}
$\theta$ 
{is}
a trainable weight initialized as $\theta=-3$. The network architecture 
{used}
to parameterize $f(\vh(t),t,\theta)$ for each experiment below are described in Appendix~\ref{appendix:experimental-details}. All experiments are conducted on a server with 2 NVIDIA Titan Xp GPUs.

\begin{table}[!ht]
\fontsize{8.0}{8.0}\selectfont
\centering
\begin{threeparttable}
\caption{The batch size and learning rate for different datasets.}\label{Tab:batch-size-learning-rate}
\begin{tabular}{cccccc}
\toprule[1.0pt]
\ \ \ Dataset\ \ \   &\ \ \ Point Cloud\ \ \  &\ \ \  MNIST\ \ \   &\ \ \  CIFAR10\ \ \  & \ \ \  Plane Vibration\ \ \  & \ \ \  Walker2D\ \ \  \cr
\midrule[0.8pt]
Batch Size  & 50 & 64 & 64 & 64 & 256 \cr
Learning Rate  & 0.01 & 0.001 & 0.001 & 0.0001 & 0.003 \cr
\bottomrule[1.0pt]
\end{tabular}
\end{threeparttable}\vspace{-0.1in}
\end{table}









\begin{figure*}[!ht]\vspace{-2mm}
\centering
\begin{tabular}{cc}
\hskip -0.3cm
\includegraphics[clip, trim=0.1cm 0.1cm 0.1cm 1.5cm,width=0.46\linewidth]{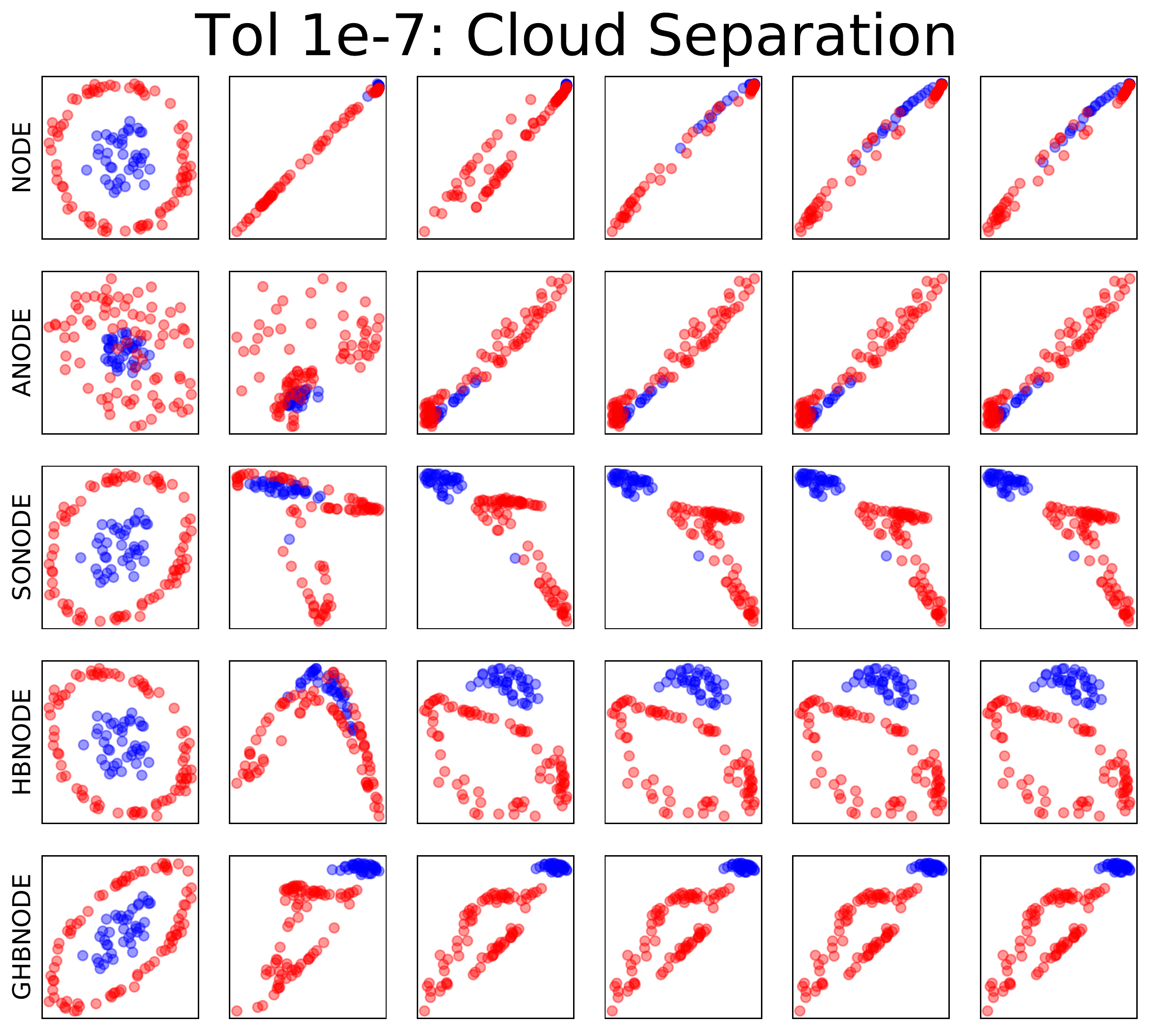}&
\hskip -0.4cm\includegraphics[clip, trim=0.1cm 0.1cm 0.1cm 0.1cm,width=0.45\linewidth]{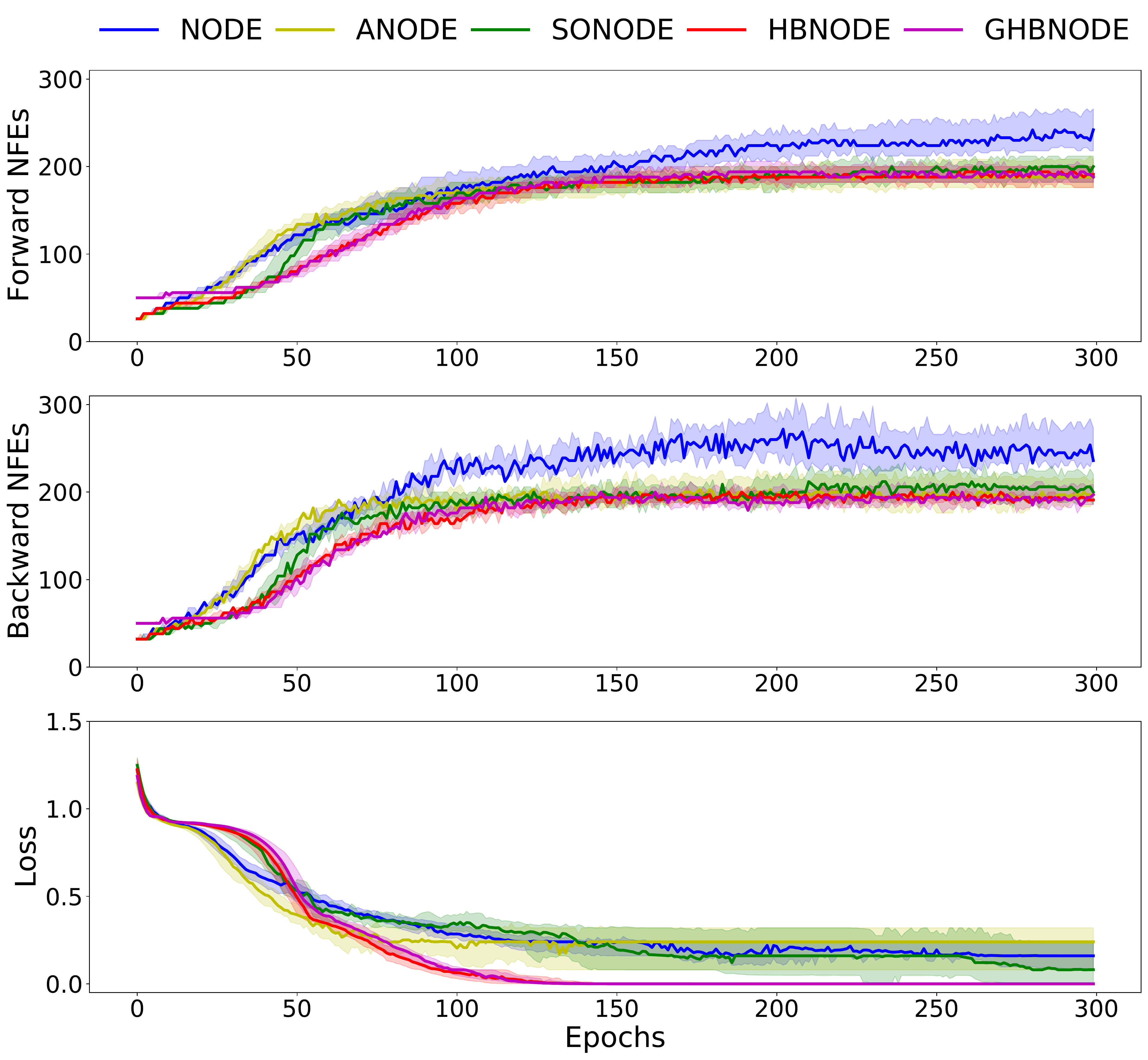}\\
\end{tabular}
\vskip -0.3cm
\caption{Comparison between NODE, ANODE, SONODE, HBNODE, and GHBNODE for two-dimensional point cloud separation. HBNODE and GHBNODE converge better and require less 
{NFEs}
in both forward and backward propagation than the other benchmark models.
}
\label{fig:point:cloud-separation}\vspace{-3mm}
\end{figure*}

\subsection{Point cloud separation}\label{sec:point-cloud}
In this subsection, we consider the two-dimensional point cloud separation benchmark.
A total of $120$ points are sampled, 
{in which}
$40$ points are drawn uniformly from the circle $\|\vr\|<0.5$, and $80$ points are drawn uniformly from the annulus $0.85<\|\vr\|<1.0$. This experiment aims to learn effective features to classify these two point clouds. 
{Following \cite{NEURIPS2019_21be9a4b},}
we use a three-layer neural network 
to parameterize the 
{right-hand side}
of each ODE-based model, 
{integrate}
the ODE-based model from $t_0=0$ to $T=1$, and
{pass the integration results to a dense layer to generate the classification results.}
We set the size 
of hidden layers so that the models have similar 
{sizes,}
and the number of 
{parameters}
of NODE, ANODE, SONODE, HBNODE, and GHBNODE are $525$, $567$, $528$, $568$, and $568$, respectively.
To avoid the effects of numerical error of the black-box ODE solver we set tolerance of ODE solver to be $10^{-7}$.
Figure~\ref{fig:point:cloud-separation} plots a randomly selected evolution of the point cloud separation for each model; we also compare the forward and backward NFEs and the training loss of these models 
(100 independent runs). 
HBNODE and GHBNODE improve training as the training loss consistently goes to zero over different runs, while ANODE and SONODE often get stuck at local 
{minima}, 
and NODE cannot separate the point cloud 
since it preserves the topology \citep{NEURIPS2019_21be9a4b}.

\subsection{Image classification}\label{sec:image-classification}



We compare the performance of HBNODE and GHBNODE with the existing ODE-based models {on} 
MNIST and CIFAR10 classification {tasks} using the same setting as 
in \cite{NEURIPS2019_21be9a4b}.
We parameterize $f(\vh(t),t,\theta)$ 
using a 3-layer convolutional network for each ODE-based model, and the total number of parameters 
{for} each model is listed in Table~\ref{Tab:num-params-image-classification}.
For a given input image of the size $c\times h\times w$, 
we first augment the number of channel from $c$ to $c+p$ 
with the augmentation dimension $p$ 
{dependent}
on each method\footnote{We set $p=0, 5, 4, 4, 5/0,10,9,9,9$ on MNIST/CIFAR10 
for NODE, ANODE, SONODE, HBNODE, and GHBNODE, respectively.}. Moreover, for SONODE, HBNODE and GHBNODE, we further include velocity or momentum with the same shape as the augmented state. 

\begin{table}[!ht]\vspace{-0.1in}
\fontsize{8.0}{8.0}\selectfont
\centering
\begin{threeparttable}
\caption{The number of parameters for each models for image classification.}\label{Tab:num-params-image-classification}
\begin{tabular}{cccccc}
\toprule[1.0pt]
\ \ \ Model\ \ \   &\ \ \ NODE\ \ \  &\ \ \  ANODE\ \ \   &\ \ \  SONODE\ \ \  &\ \ \  HBNODE\ \ \  &\ \ \  GHBNODE\ \ \  \cr
\midrule[0.8pt]
\#Params (MNIST)    & 85,315 & 85,462 & 86,179 & {85,931} & {85,235} \cr
\#Params (CIFAR10)  & 173,611 & 172,452 & 171,635 & 172,916 & 172,916 \cr
\bottomrule[1.0pt]
\end{tabular}
\end{threeparttable}\vspace{-0.2in}
\end{table}

\begin{figure}
\begin{center}
\begin{tabular}{cccc}
\hskip -0.3cm
\includegraphics[width=0.2\columnwidth]{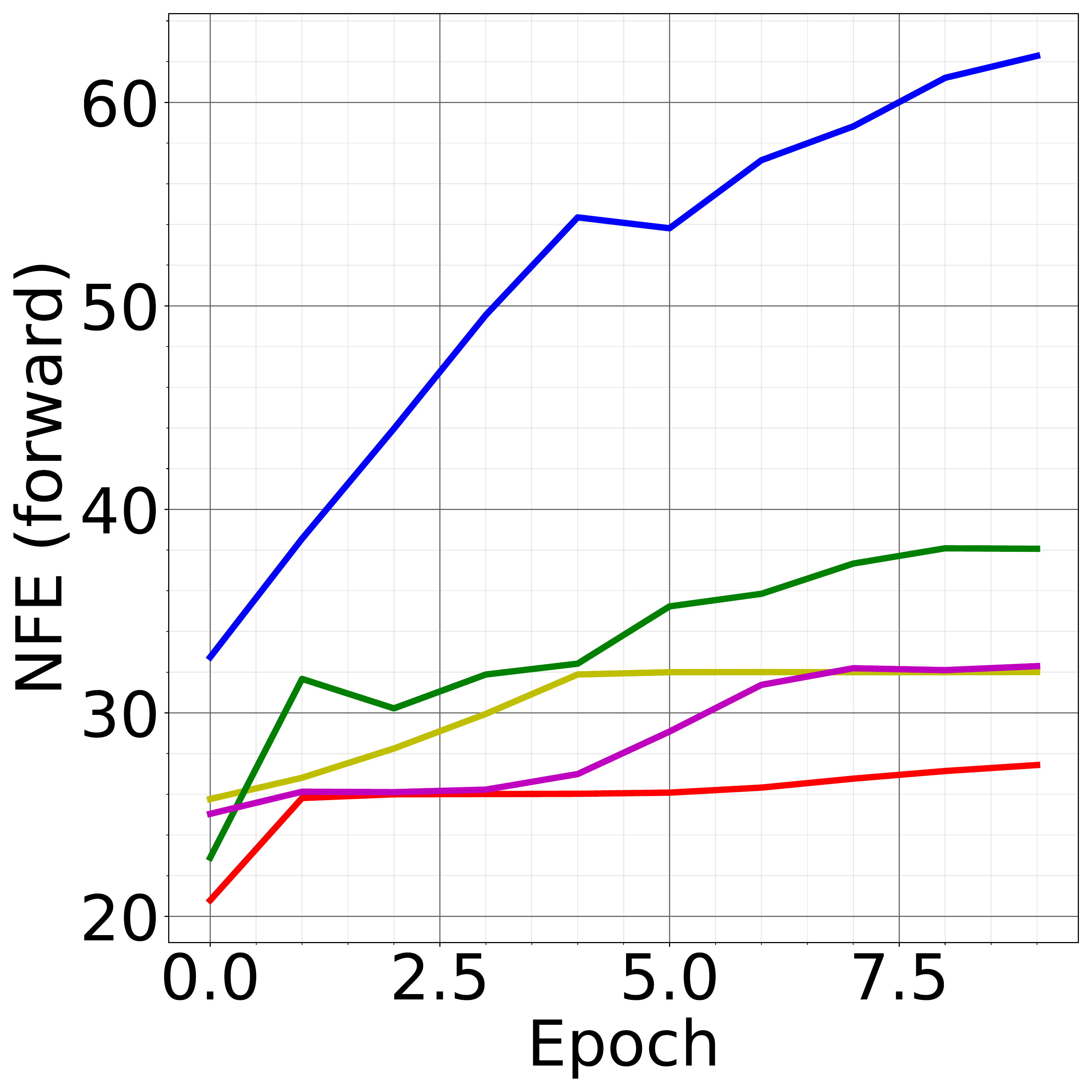}&
\hskip -0.3cm
\includegraphics[width=0.2\columnwidth]{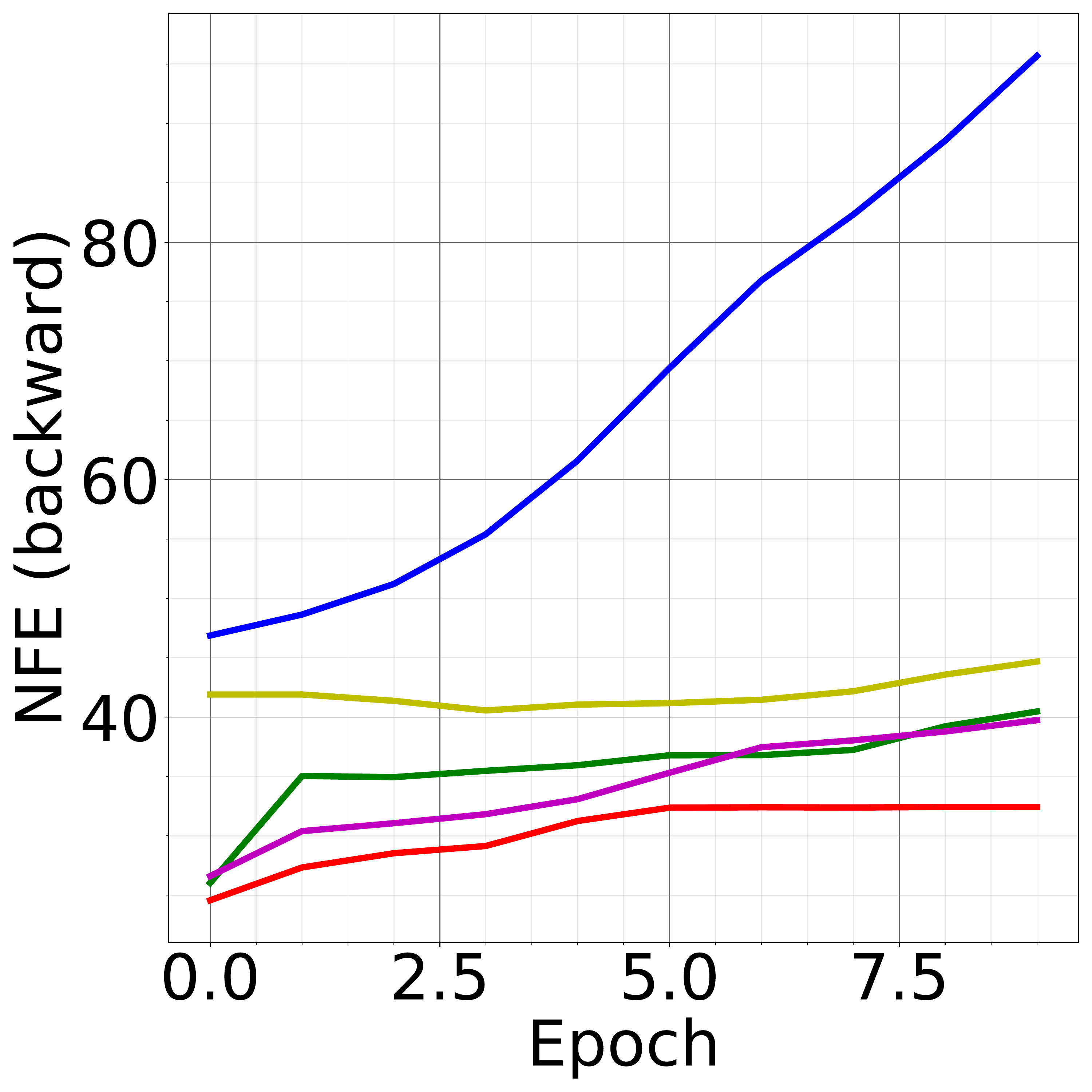}& 
\hskip -0.3cm
\includegraphics[width=0.2\columnwidth]{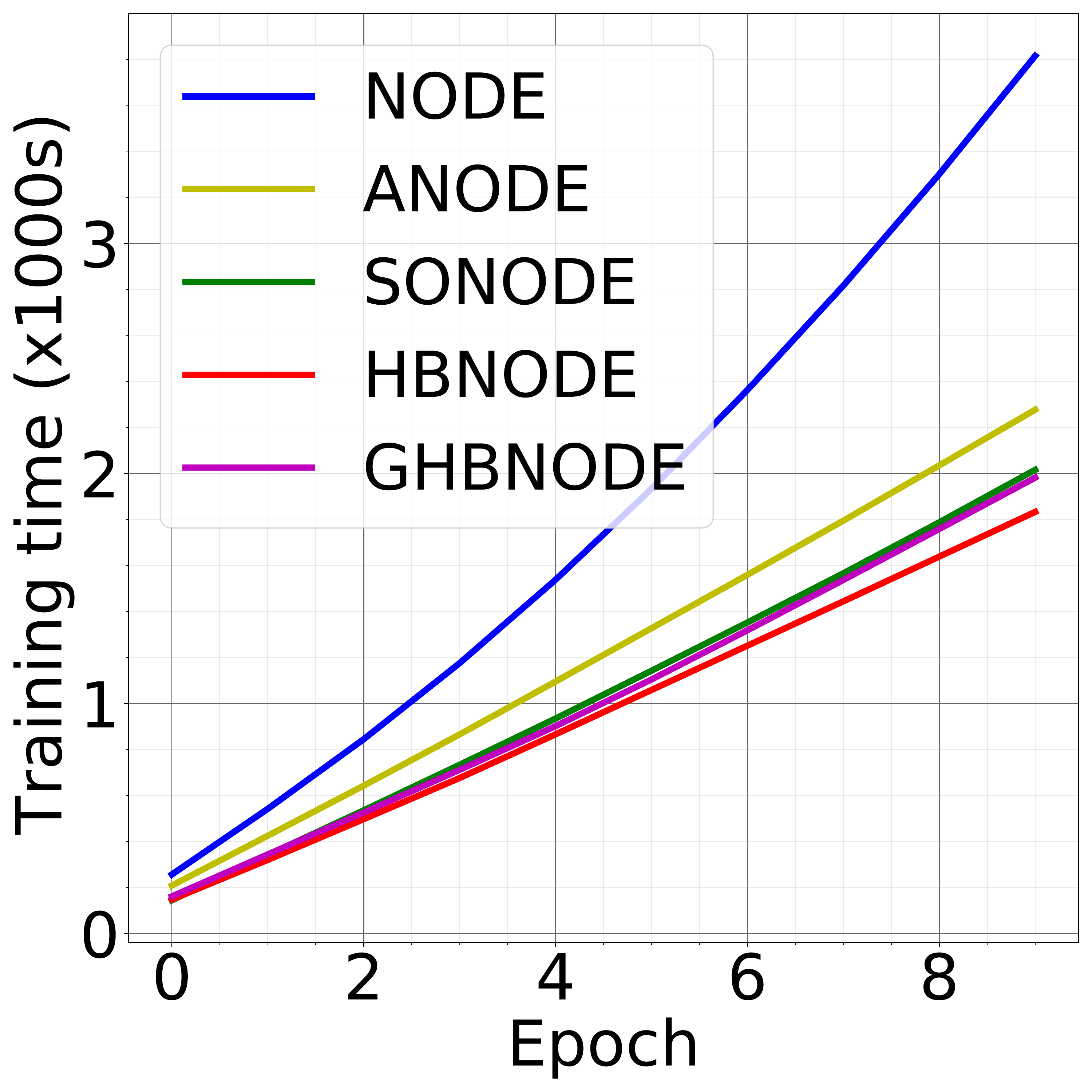} &
\hskip -0.3cm
\includegraphics[width=0.195\columnwidth]{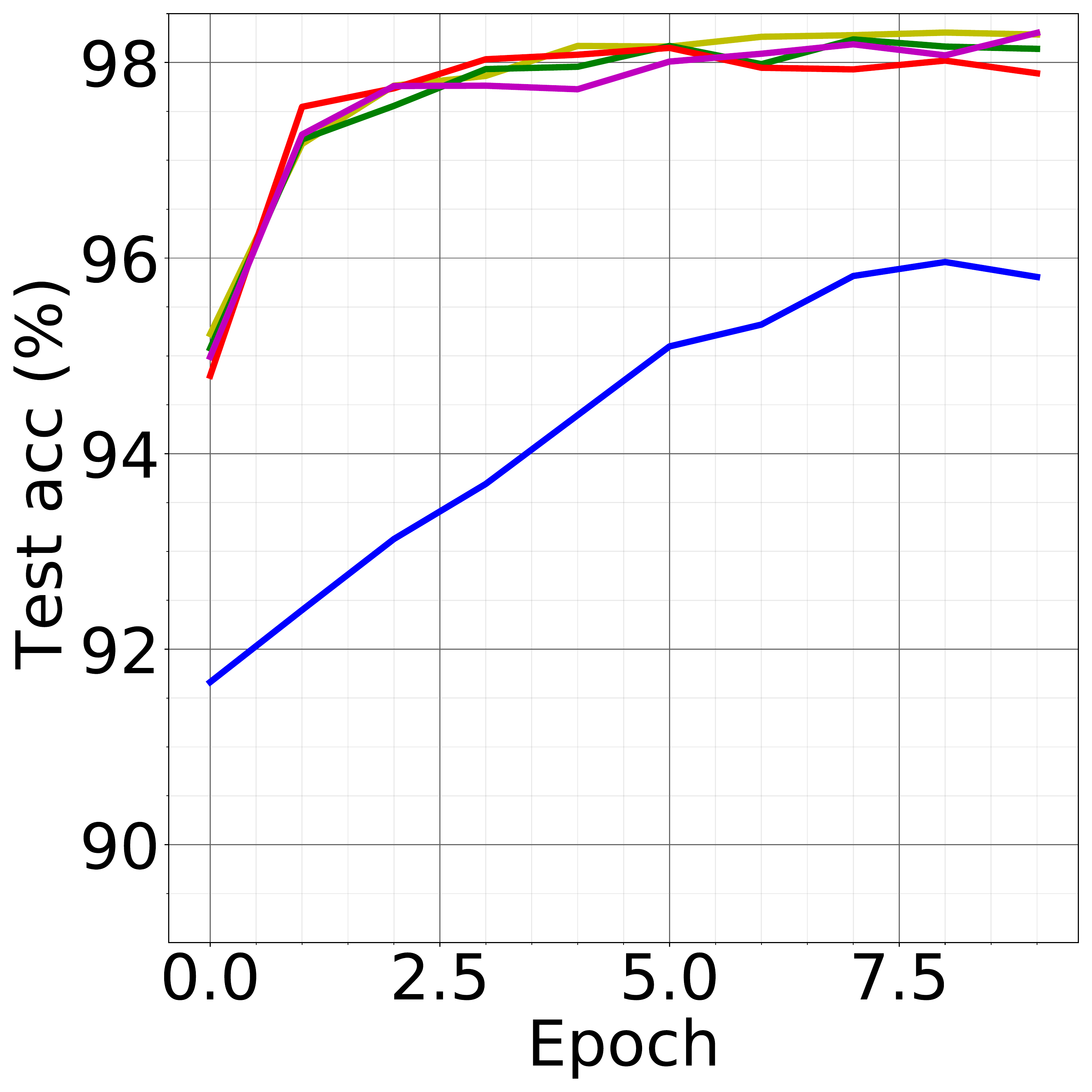}\\
  \end{tabular}
  \end{center}
  \vskip -0.2in
  \caption{ Contrasting NODE \citep{chen2018neural}, ANODE \citep{NEURIPS2019_21be9a4b}, SONODE \citep{norcliffe2020_sonode}, HBNODE, and GHBNODE for MNIST classification in NFE, training time, and test accuracy. (Tolerance: $10^{-5}$).  
  }\label{fig:node-hbnode-mnist}
\end{figure}

\paragraph{NFEs.}
As shown in Figs.~\ref{fig:node-hbnode-cifar10} and \ref{fig:node-hbnode-mnist}, the NFEs grow rapidly with training of the NODE, resulting in an increasingly 
complex model with reduced performance and the possibility 
of blow up. Input augmentation has been verified to effectively reduce the NFEs, as 
both ANODE and SONODE 
{require fewer}
forward 
{NFEs}
than NODE for the MNIST and CIFAR10 classification. However, input augmentation is less effective {in controlling} 
their backward NFEs. HBNODE and GHBNODE require much 
{fewer}
NFEs than the existing benchmarks, especially for backward NFEs. In practice, reducing NFEs {implies reducing} 
both training and inference time, as shown in Figs.~\ref{fig:node-hbnode-cifar10} and \ref{fig:node-hbnode-mnist}. 



\paragraph{Accuracy.}
We {also} compare the accuracy of different ODE-based models for MNIST and CIFAR10 classification. As shown in
{Figs.~\ref{fig:node-hbnode-cifar10} and \ref{fig:node-hbnode-mnist}}, HBNODE and GHBNODE have slightly better classification accuracy than the other three models; this resonates with the fact that less NFEs 
{lead}
to simpler models which generalize better \citep{NEURIPS2019_21be9a4b,norcliffe2020_sonode}. 

\begin{figure}[!ht]\vspace{-2mm}
\centering
\begin{tabular}{c}
\includegraphics[clip, trim=0.01cm 0.01cm 0.01cm 0.01cm, width=0.99\columnwidth]{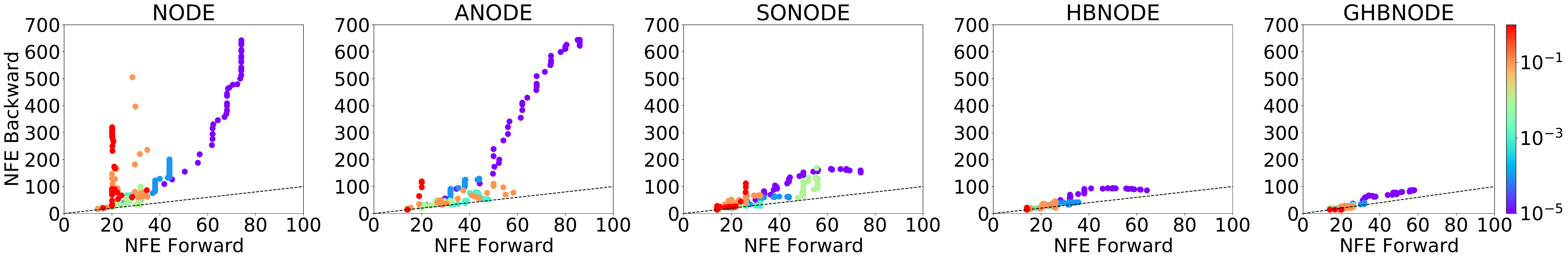}\\
\end{tabular}
\vspace{-4mm}
\caption{NFE vs. tolerance (shown in the colorbar) for training ODE-based models for CIFAR10 classification. Both forward and backward NFEs of HBNODE and GHBNODE grow much 
{more slowly}
than that of NODE, ANODE, and SONODE; especially the backward 
{NFEs}.
As the tolerance decreases, the advantage of HBNODE and GHBNODE in reducing NFEs becomes more significant.}
\label{fig:NFE-vs-tol}\vspace{-5mm}
\end{figure}

\paragraph{NFEs vs. tolerance.} We further study the NFEs for different ODE-based models under different 
{tolerances}
of the 
ODE solver using the same approach as 
in \cite{chen2018neural}. Figure~\ref{fig:NFE-vs-tol} depicts the forward and backward NFEs for different models under different tolerances. We see that (i) both forward and backward NFEs grow quickly when tolerance 
{is decreased,}
and HBNODE and GHBNODE require much 
fewer NFEs than other models; (ii) under different tolerances, the backward
{NFEs}
of NODE, ANODE, and SONODE 
{are much larger}
than the forward 
{NFEs,}
and the difference becomes 
larger when 
{the}
tolerance decreases. In contrast, the forward and backward NFEs of HBNODE and GHBNODE scale almost linearly with each other. 
{This} reflects that the advantage in NFEs of (G)HBNODE over {the} benchmarks become more significant when a smaller tolerance is used.

\subsection{Learning dynamical systems from irregularly-sampled time series}\label{sec:physical-systems}
In this subsection, we learn dynamical systems from experimental measurements. In particular, we use the ODE-RNN framework \citep{chen2018neural,NEURIPS2019_42a6845a}, with the recognition model being set to different ODE-based models, 
to study the vibration of an airplane dataset \citep{noel2017f}. The dataset was acquired, from time $0$ to $73627$, by attaching a shaker underneath the right wing to provide input signals, and $5$ attributes are recorded per time stamp; these attributes include voltage of input signal, force applied to aircraft, and acceleration at $3$ different spots of the airplane. We randomly take out $10\%$ of the data to make the time series irregularly-sampled. We use the first $50\%$ of data as our train set, the next $25\%$ as validation set, and the rest as test set. 
We divide each set into non-overlapping segments of consecutive $65$ time stamps of the irregularly-sampled time series, with each input instance consisting of $64$ time stamps of the irregularly-sampled time series, and we aim to forecast $8$ consecutive time stamps starting from the last time stamp of the segment. The input is fed through the the hybrid methods in a recurrent fashion; by changing the time duration of the last step of the ODE integration, we can forecast the output in the different time stamps. The output of the hybrid method is passed to a single dense layer to generate the output time series. In our experiments,  we compare different ODE-based models hybrid with RNNs. The 
ODE of each model is parametrized by a $3$-layer network whereas the RNN is parametrized by a simple dense network; the total number of parameters for ODE-RNN, ANODE-RNN, SONODE-RNN, HBNODE-RNN, and GHBNODE-RNN with $16$, $22$, $14$, $15$, $15$ augmented dimensions are 15,986, 16,730, 16,649, 16,127, and 16,127, respectively. To avoid potential error due to the ODE solver, we use a tolerance of $10^{-7}$.



In training those hybrid models, we regularize the models by penalizing the L2 distance between the RNN output and the values of the next time stamp.
Due to the second-order natural of the underlying dynamics \citep{norcliffe2020_sonode}, ODE-RNN and ANODE-RNN learn the dynamics very poorly with much larger training and test losses than the other models even they take smaller NFEs. HBNODE-RNN and GHBNODE-RNN give better prediction than SONODE-RNN using less backward NFEs.

\begin{figure*}[!ht]
\centering
\begin{tabular}{ccccc}
\hskip -0.4cm
\includegraphics[width=0.2\linewidth]{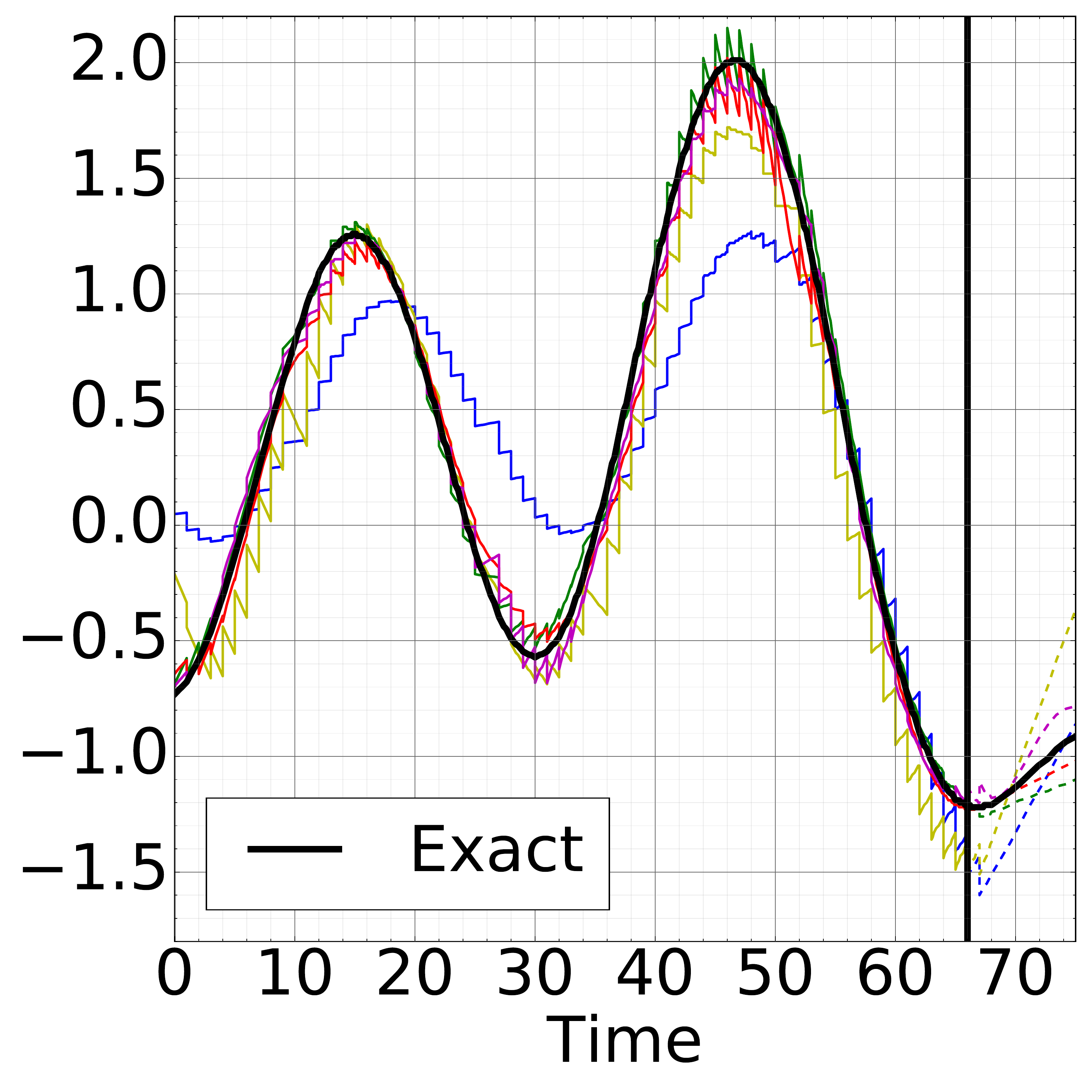}&
\hskip -0.4cm
\includegraphics[width=0.2\linewidth]{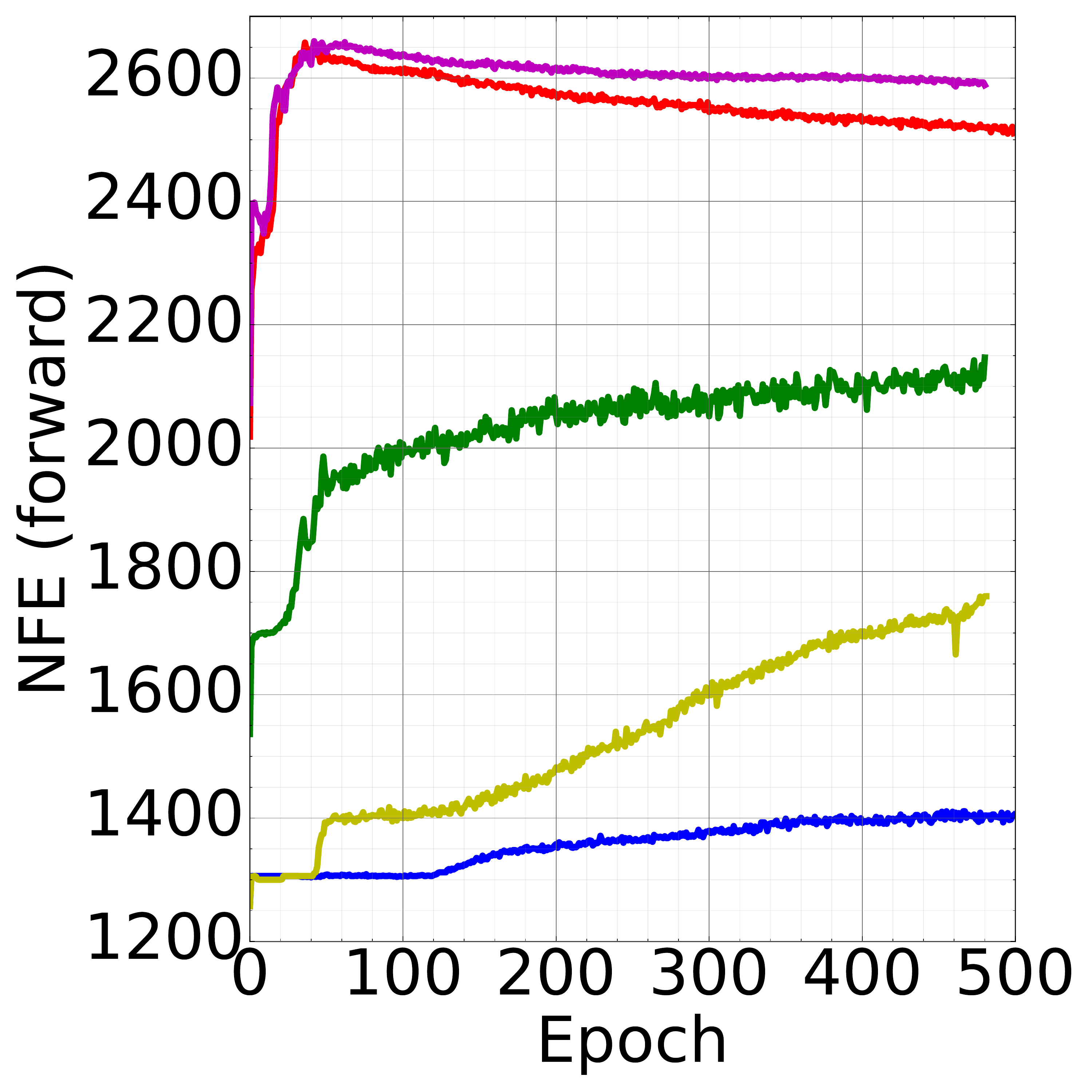}&
\hskip -0.4cm
\includegraphics[width=0.2\linewidth]{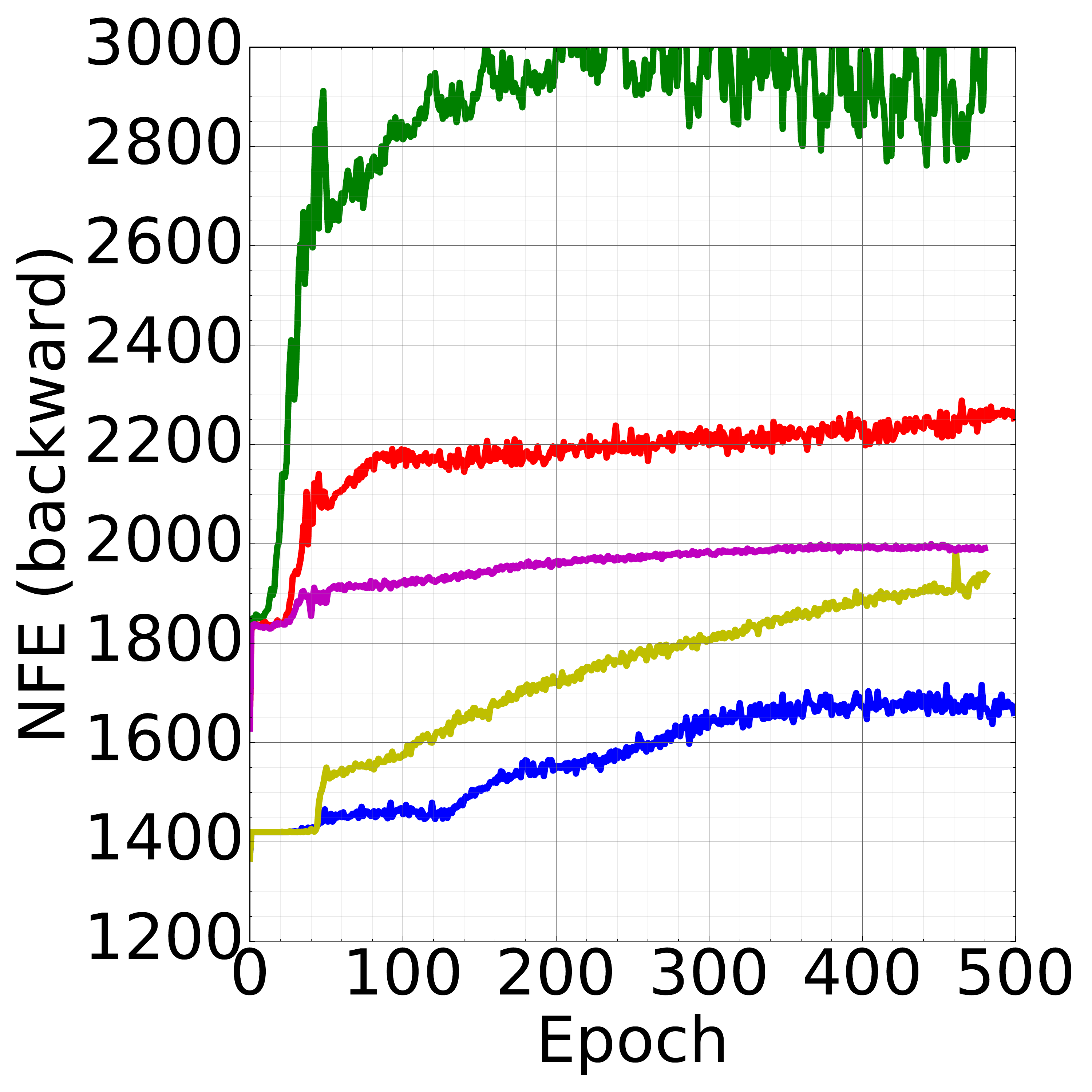}&
\hskip -0.4cm
\includegraphics[width=0.2\linewidth]{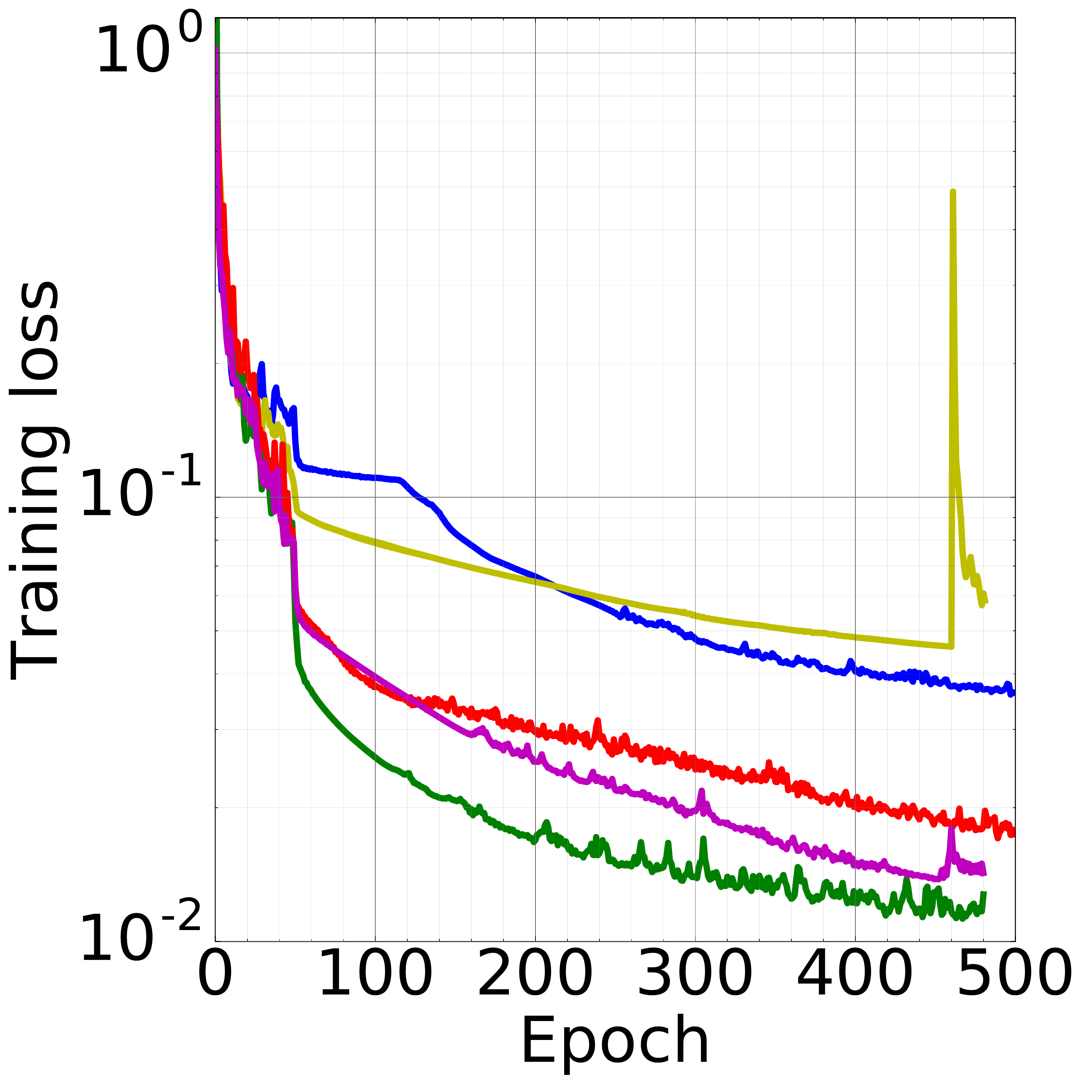}&
\hskip -0.4cm
\includegraphics[width=0.2\linewidth]{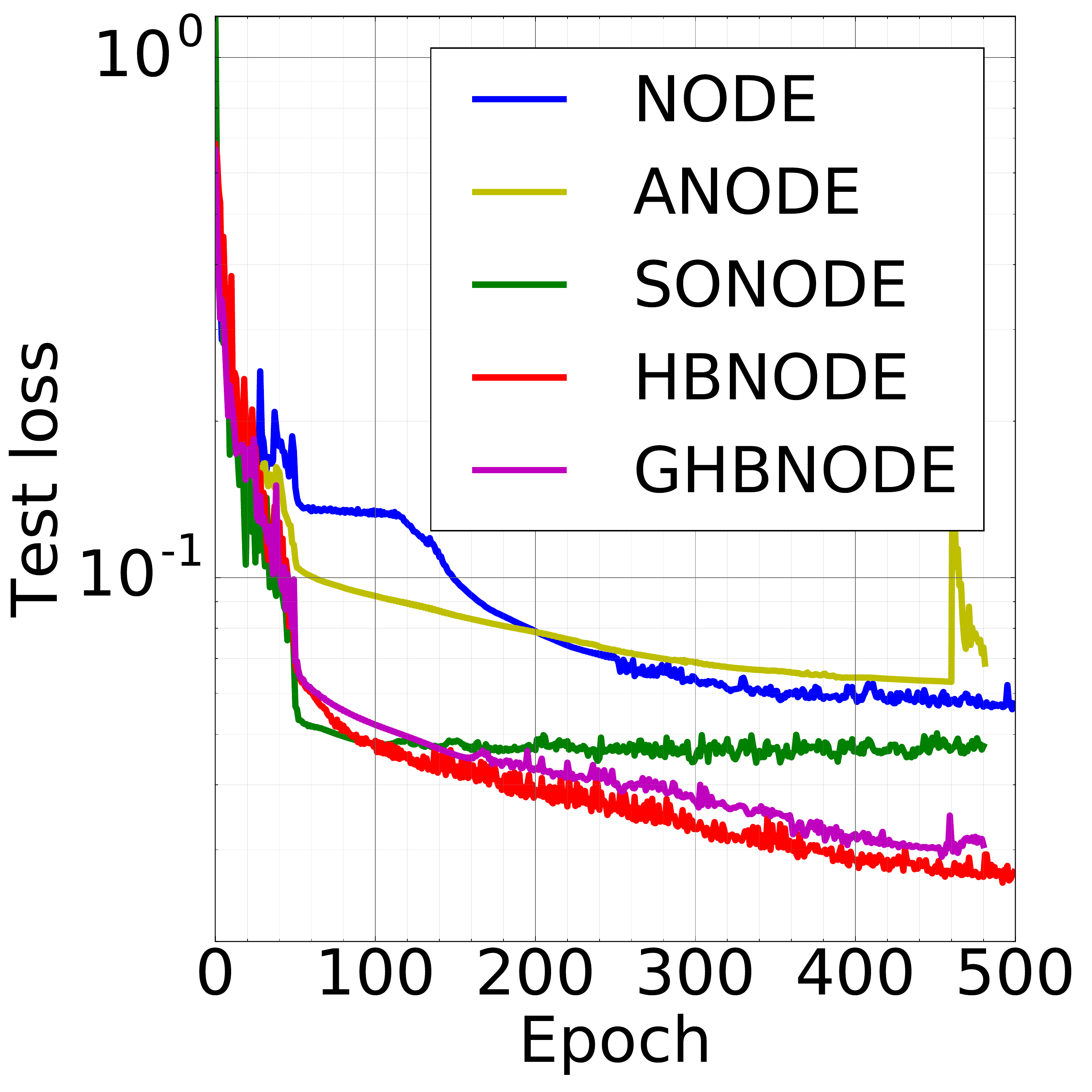}\\
\end{tabular}
\vskip -0.3cm
\caption{Contrasting ODE-RNN, ANODE-RNN, SONODE-RNN, HBNODE-RNN, and GHBNODE-RNN for learning a vibrational dynamical system. Left most: The learned curves of each model vs. the ground truth (Time: $<$66 for training, 66-75 for testing).  
}
\label{fig:Vibration-dynamics}
\end{figure*}

\subsection{
Walker2D kinematic simulation
}\label{sec:sequential-modeling}

\begin{figure*}[!ht]
\centering
\begin{tabular}{ccccc}
\hskip -0.3cm
\includegraphics[width=0.2\linewidth]{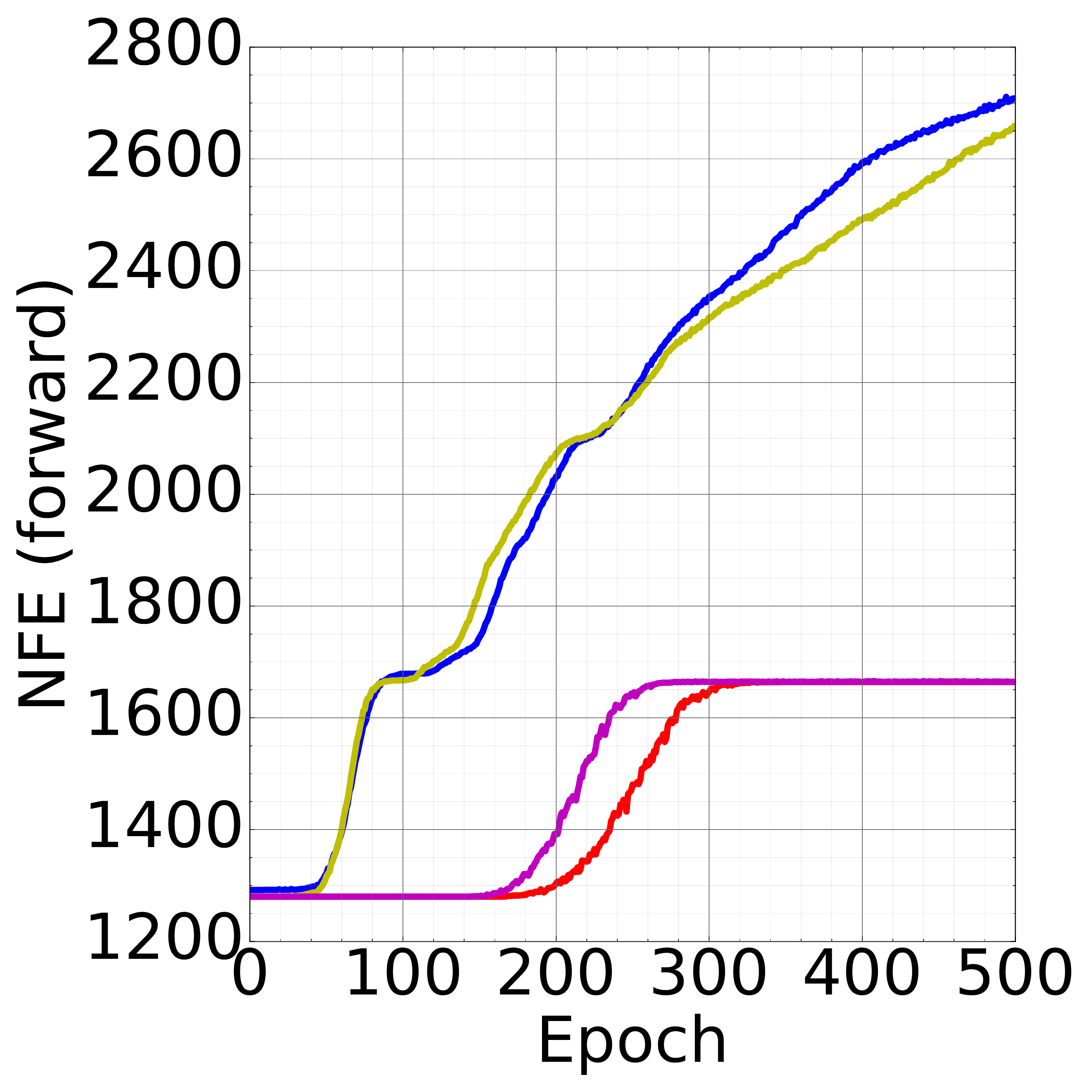}&
\hskip -0.4cm
\includegraphics[width=0.2\linewidth]{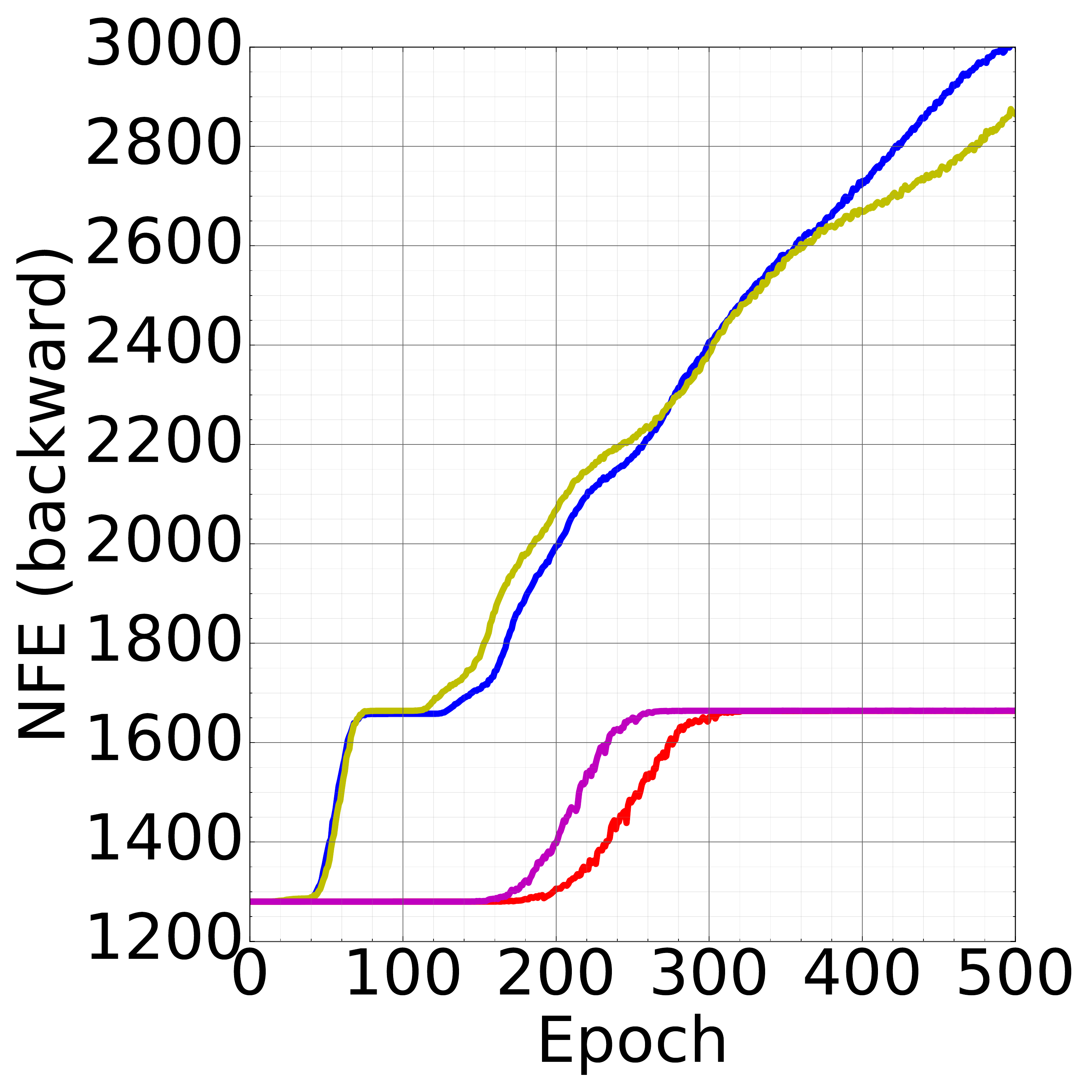}&
\hskip -0.4cm
\includegraphics[width=0.2\linewidth]{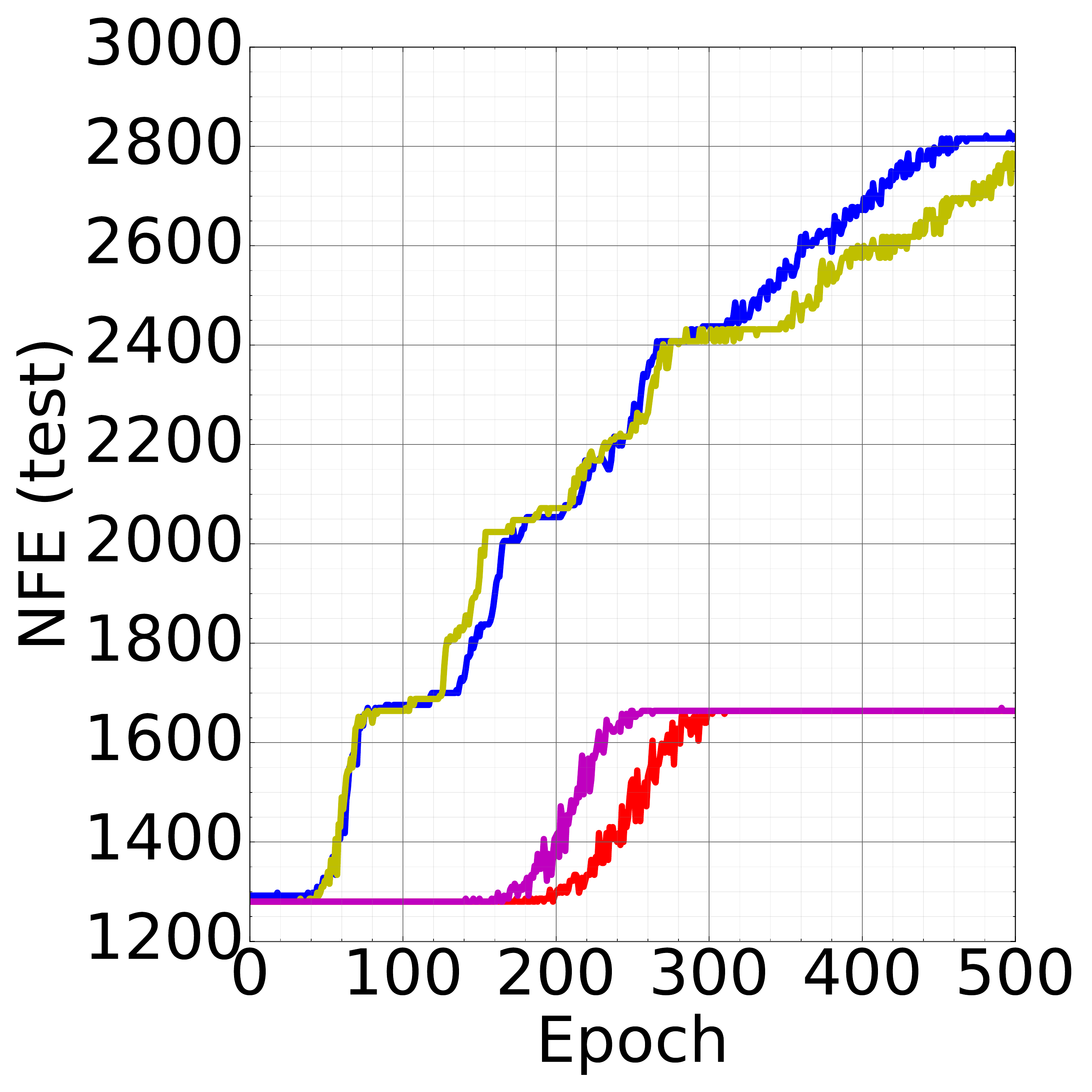}&
\hskip -0.4cm
\includegraphics[width=0.2\linewidth]{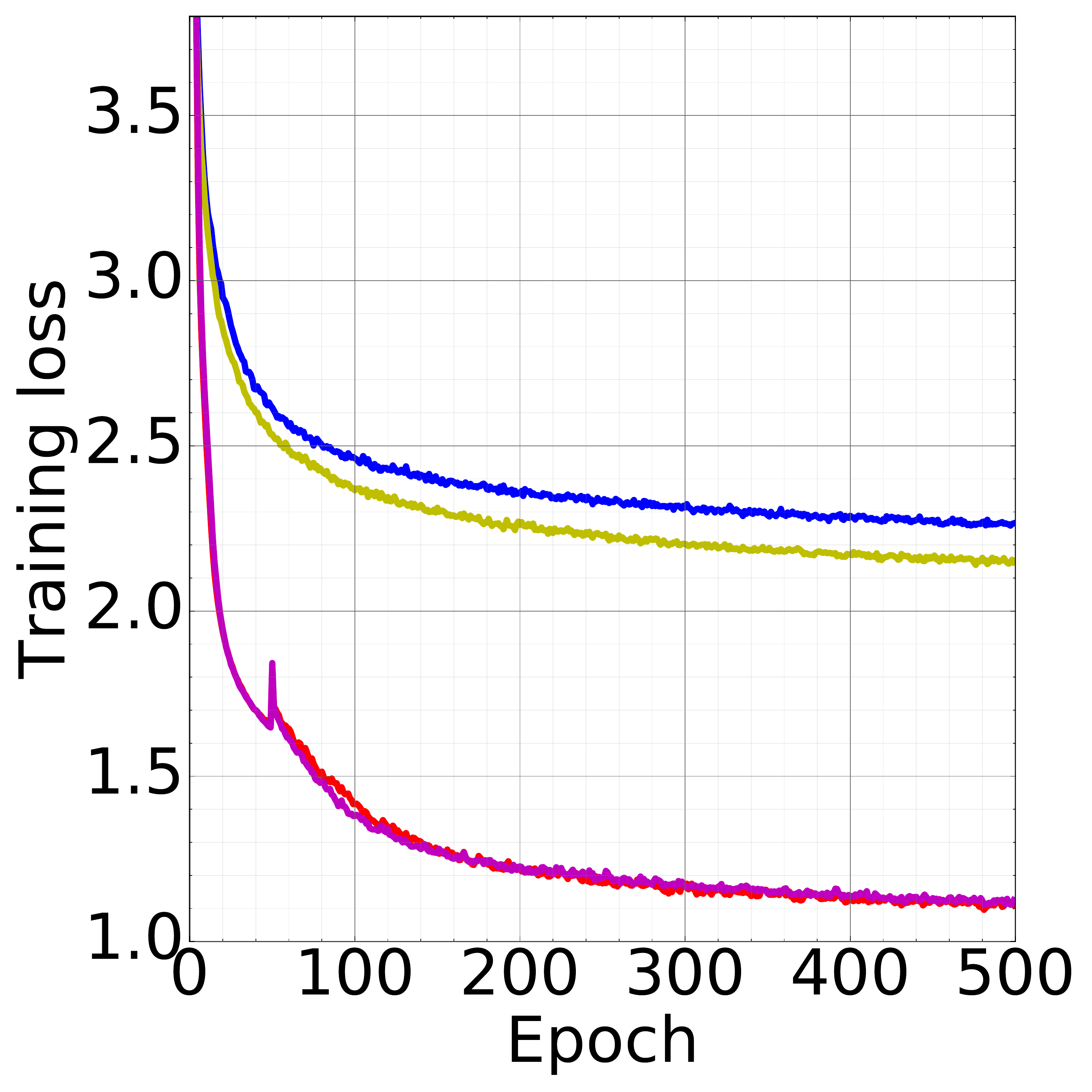}&
\hskip -0.4cm
\includegraphics[width=0.2\linewidth]{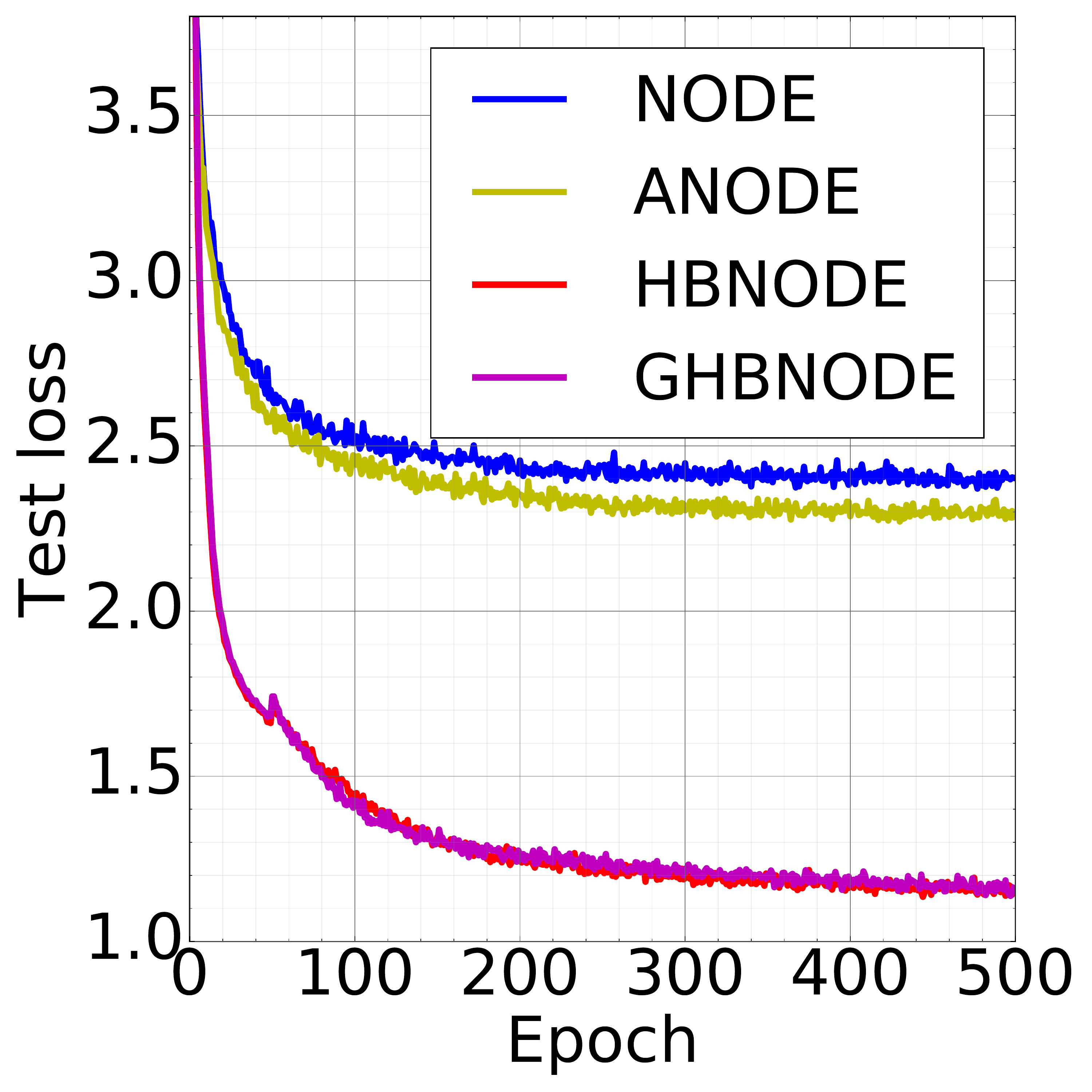}\\
\end{tabular}
\vskip -0.3cm
\caption{Contrasting ODE-RNN, ANODE-RNN, SONODE-RNN, HBNODE-RNN, and GHBNODE-RNN for the Walker-2D kinematic simulation. 
}
\label{fig:walker-dynamics}
\end{figure*}

In this subsection, we evaluate the performance of HBNODE-RNN and GHBNODE-RNN on the 
Walker2D kinematic simulation task, which requires learning long-term dependency effectively \citep{lechner2020learning}. The dataset \citep{brockman2016openai} consists of a dynamical system from kinematic simulation of a person walking from a pre-trained policy, aiming to learn the kinematic simulation of the MuJoCo physics engine \citep{6386109}. The dataset is irregularly-sampled where $10\%$ of data are removed from the simulation. Each input is consisted of 64 time stamps and fed though the the hybrid methods in a recurrent fashion, and the outputs of hybrid methods is passed to a single dense layer to generate the output time series. The target is to provide auto-regressive forecast so that the output time series is as close as the input sequence shifted 1 time stamp to the right. We compare ODE-RNN 
(with 7 augmentation), ANODE-RNN (with 7 ANODE style augmentation), HBNODE-RNN 
(with 7 augmentation), and GHBNODE-RNN (with 7 augmentation) \footnote{Here, we do not compare with SONODE-RNN since SONODE has some initialization problem on this dataset, and the ODE solver encounters failure due to exponential growth over time. This issue is originally tackled by re-initialization \citep{norcliffe2020_sonode}. 
We re-initialized SONODE 100 times; all failed 
due to initialisation problems.}. The RNN is parametrized by a 3-layer network whereas the ODE is parametrized by a simple dense network. The number of parameters of the above four models are 8,729, 8,815, 8,899, and 8,899, respectively.  In Fig.~\ref{fig:walker-dynamics}, we compare the performance of the above four models on the Walker2D benchmark; HBNODE-RNN and GHBNODE-RNN not only require significantly less NFEs in both training (forward and backward) and in testing than ODE-RNN and ANODE-RNN, but also have much smaller training and test losses. 

\section{Related Work}\label{sec:related-works}
\vspace{-0.05in}\paragraph{Reducing NFEs in training NODEs.} Several techniques have been developed to reduce the NFEs for the forward solvers in NODEs, including weight decay \citep{grathwohl2018scalable}, input augmentation \citep{NEURIPS2019_21be9a4b},  
regularizing solvers and learning dynamics 
\citep{pmlr-v119-finlay20a,NEURIPS2020_2e255d2d,NEURIPS2020_a9e18cb5,pmlr-v139-pal21a},
high-order ODE \citep{norcliffe2020_sonode}, data control \citep{massaroli2020dissecting}, and depth-variance \citep{massaroli2020dissecting}. HBNODEs can 
{reduce}
both forward and backward NFEs at the same time.

\vspace{-0.12in}\paragraph{Second-order ODE accelerated dynamics.} It has been noticed in both optimization and sampling communities that second-order ODEs with an appropriate damping term, e.g., the classical momentum and Nesterov's acceleration in discrete regime, can significantly accelerate the first-order gradient dynamics (gradient descent), e.g., \citep{polyak1964some,nesterov1983method,
chen2014stochastic,su2014differential,
wilson2018lyapunov}. Also, these second-order ODEs have 
been discretized via some interesting numerical schemes to design fast optimization schemes, e.g., \cite{NEURIPS2019_a9986cb0}.




\vspace{-0.12in}\paragraph{Learning long-term dependencies.} Learning long-term dependency is one of the most important goals for learning from sequential data. Most of the existing works 
{focus on}
mitigating exploding or vanishing gradient issues in training RNNs, e.g., \citep{arjovsky2016unitary,wisdom2016full,jing2017tunable,vorontsov2017orthogonality,mhammedi2017efficient,pmlr-v80-helfrich18a,MomentumRNN}. Attention-based models are proposed for learning on sequential data concurrently with the effective accommodation of learning long-term dependency \citep{NIPS2017_3f5ee243,devlin2018bert}. Recently, NODEs have been integrated with long-short term memory model \citep{hochreiter1997long} to learn long-term dependency for irregularly-sampled time series \citep{lechner2020learning}. HBNODEs directly enhance learning long-term dependency from sequential data. 

\vspace{-0.12in}\paragraph{
Momentum in neural network design.} As a line of orthogonal work, the momentum has also been studied in designing neural network architecture, e.g., \citep{
moreau2017understanding,
MomentumRNN,li2018optimization,sander2021momentum}, 
which can also help accelerate training and learn long-term dependencies. These techniques can be considered as changing the neural network $f$ in \eqref{eq:NODE}. 
{We} leave the synergistic integration of adding momentum to $f$ with our work on changing the 
{left-hand side}
of \eqref{eq:NODE} as a future work. 

\vspace{-0.12in}\paragraph{ResNet-style models.
} Interpreting ResNet as an ODE model has been an interesting 
research direction 
which has lead to interesting neural network architectures and analysis from the numerical ODE solvers and differential equation theory viewpoints, e.g., \cite{lu2018beyond,li2019deep,doi:10.1137/19M1265302}.

\section{Concluding Remarks}\label{sec:conclusion}
We proposed HBNODEs to 
reduce the NFEs in 
solving both forward and backward ODEs,  
which also improve generalization performance over the existing benchmark models. Moreover, HBNODEs alleviate vanishing gradients in training NODEs, making HBNODEs able to learn long-term dependency effectively from 
sequential data. In the optimization community, Nesterov acceleration \citep{nesterov1983method} is also a famous algorithm for accelerating gradient descent, that achieves an optimal convergence rate for general convex optimization problems. The ODE counterpart of the Nesterov's acceleration corresponds to \eqref{eq:HBNODE-2nd} with $\gamma$ being replaced by a time-dependent damping parameter, e.g., $t/3$ \citep{su2014differential} or with restart \cite{wang2020scheduled}. The adjoint equation of the Nesterov's ODE \citep{su2014differential} is no longer a Nesterov's ODE. We notice that directly using Nesterov's ODE cannot improve the performance of the vanilla neural ODE. How to integrate Nesterov's ODE with neural ODE is an interesting future direction. Another interesting direction is connecting HBNODE with symplectic ODE-net \citep{Zhong2020Symplectic} through an appropriate change of variables.

\section{Acknowledgement}
This material is based on research sponsored by the NSF grant DMS-1924935 and DMS-1952339, the DOE grant  DE-SC0021142, and the ONR grant N00014-18-1-2527 and ONR MURI grant N00014-20-1-2787.

\clearpage
\bibliographystyle{plainnat}

\clearpage
\section*{Checklist}
\begin{enumerate}

\item For all authors...
\begin{enumerate}
  \item Do the main claims made in the abstract and introduction accurately reflect the paper's contributions and scope?
    \answerYes{}
  \item Did you describe the limitations of your work?
    \answerYes{See Section 4.1.}
  \item Did you discuss any potential negative societal impacts of your work?
    \answerNA{}
  \item Have you read the ethics review guidelines and ensured that your paper conforms to them?
    \answerYes{}
\end{enumerate}

\item If you are including theoretical results...
\begin{enumerate}
  \item Did you state the full set of assumptions of all theoretical results?
    \answerYes{See Section 4.1.}
	\item Did you include complete proofs of all theoretical results?
    \answerYes{ See   Supplementary Materials}
\end{enumerate}

\item If you ran experiments...
\begin{enumerate}
  \item Did you include the code, data, and instructions needed to reproduce the main experimental results (either in the supplemental material or as a URL)?
    \answerYes{}
  \item Did you specify all the training details (e.g., data splits, hyperparameters, how they were chosen)?
    \answerYes{}
	\item Did you report error bars (e.g., with respect to the random seed after running experiments multiple times)?
    \answerYes{}
	\item Did you include the total amount of compute and the type of resources used (e.g., type of GPUs, internal cluster, or cloud provider)?
    \answerYes{}
\end{enumerate}

\item If you are using existing assets (e.g., code, data, models) or curating/releasing new assets...
\begin{enumerate}
  \item If your work uses existing assets, did you cite the creators?
    \answerYes{}
  \item Did you mention the license of the assets?
    \answerYes{}
  \item Did you include any new assets either in the supplemental material or as a URL?
    \answerNA{}
  \item Did you discuss whether and how consent was obtained from people whose data you're using/curating?
    \answerNA{}
  \item Did you discuss whether the data you are using/curating contains personally identifiable information or offensive content?
    \answerNA{}
\end{enumerate}

\item If you used crowdsourcing or conducted research with human subjects...
\begin{enumerate}
  \item Did you include the full text of instructions given to participants and screenshots, if applicable?
    \answerNA{}
  \item Did you describe any potential participant risks, with links to Institutional Review Board (IRB) approvals, if applicable?
    \answerNA{}
  \item Did you include the estimated hourly wage paid to participants and the total amount spent on participant compensation?
    \answerNA{}
\end{enumerate}

\end{enumerate}

\clearpage
\appendix
\begin{center}
{\Large \bf Supplementary Material for\\ \textit{Heavy Ball Neural Ordinary Differential Equations}}
\end{center}

\section{Review of the Adjoint Equation for the First- and Second-order ODEs}

The adjoint sensitivity method is the key to assuring constant memory usage in training neural ODEs \citep{chen2018neural}. In this section, we present two different proofs for 
the first-order adjoint sensitivity equations. The differentiation proof in 
Appendix \ref{differentiation-proof} is adapted from the proof by \cite{norcliffe2020_sonode}. We provide a new integral proof in 
Appendix \ref{integraion-proof} to extend theoretical support for the Lipschitz continuous functions. We also revisit the proof of the second-order adjoint sensitivity equations by \cite{norcliffe2020_sonode}.

\subsection{First-order Adjoint Sensitivity Equation}
\label{1st-order-proofs}
A Neural ODE 
{for hidden features $\vh(t)\in \RR^N$ takes the form}
\begin{equation}
\frac{\partial \vh}{\partial t} = f(\vh(t), t, \theta),
\quad 
\vh(t_0) = \vh_{t_0},
\quad
\vh(T) = \vh_T,
\end{equation}
{where $f(\vh(t),t,\theta)\in \RR^N$ is a neural network with learnable parameters $\theta$.}
The corresponding adjoint equation, with $\mathcal{L}$ 
{being} 
a scalar loss function,
is 
{defined by the following ODE,} 
\begin{equation}\label{eq:Appendix-NODE-Adjoint}
\frac{\partial \mA(t)}{\partial t} = -\mA(t) \frac{\partial f}{\partial \vh},
\quad 
\mA(T) = -\mI,
\quad 
\va(t) = -\frac{d\mathcal{L}}{d\vh_T} \mA(t).
\end{equation}
For gradient-based optimization, we need to compute the following derivatives
\begin{equation}
\frac{d\mathcal{L}}{d \theta} = \frac{d\mathcal{L}}{d \vh_T}\frac{d \vh_T}{d \theta},
\quad 
\frac{d\mathcal{L}}{d \vh_{t_0}} = \frac{d\mathcal{L}}{d \vh_T}\frac{d \vh_T}{d \vh_{t_0}}.
\end{equation}
In the following sections, we show that
\begin{equation}
\frac{d \vh_T}{d \theta} = \int_{T}^{t_0}\mA \frac{\partial f}{\partial \theta} dt,
\quad 
\frac{d \vh_T}{d \vh_{t_0}} = -\mA(t_0).
\label{eq:1st-order-A}
\end{equation}
By linearity, we immediately arrive at the following adjoint sensitivity equations 
\begin{equation}
\frac{d \mathcal{L}}{d \theta} = \int_{t_0}^{T}\va \frac{\partial f}{\partial \theta} dt,
\quad 
\frac{d \mathcal{L}}{d \vh_{t_0}} = \va (t_0).
\end{equation}

\subsubsection{Proof of the First-Order Adjoint Sensitivity Equation: Differentiation Approach}\label{differentiation-proof}

We adapt the proof of the adjoint sensitivity equations from \cite{norcliffe2020_sonode}. Assume that \(f\in C^1\), $\phi$ is either $\theta$ or $\vh_{t_0}$,
then the following equations hold
\begin{equation}\label{eq:A11_1}
\frac{\partial \mA(t)}{\partial t} = -\mA(t) \frac{\partial f}{\partial \vh},
\quad
\frac{\partial^2 \vh}{\partial \phi \partial t} = \frac{\partial f}{\partial \theta}\frac{d\theta}{d\phi} + \frac{\partial f}{\partial \vh}\frac{d\vh}{d\phi},
\quad
\frac{\partial \big(\mA\frac{\partial \vh}{\partial \phi}\big)}{\partial t} = \frac{\partial \mA}{\partial t}\frac{\partial \vh}{\partial \phi} + \mA\frac{\partial^2 \vh}{\partial \phi\partial t}.
\end{equation}

{Combining the three equations in \eqref{eq:A11_1}}
yields the 
differential equation
\begin{equation}\label{eq:derivative-3eqs}
\frac{\partial \big(\mA\frac{\partial \vh}{\partial \phi}\big)}{\partial t} = \frac{\partial \mA}{\partial t}\frac{\partial \vh}{\partial \phi} + \mA\frac{\partial^2 \vh}{\partial \phi\partial t} = -\mA(t) \frac{\partial f}{\partial \vh}\frac{\partial \vh}{\partial \phi} + \mA\Big(\frac{\partial f}{\partial \theta}\frac{d\theta}{d\phi} + \frac{\partial f}{\partial \vh}\frac{d\vh}{d\phi}\Big) = \mA\frac{\partial f}{\partial \theta}\frac{d\theta}{d\phi}.
\end{equation}
{Integrating}
both sides of \eqref{eq:derivative-3eqs} in $t$ from $T$ to $t_0$, we arrive at the 
integral equation
\begin{equation}\label{eq:A113}
\Big(\mA\frac{\partial \vh}{\partial \phi}\Big)\Big|_{T}^{t_0} = \int_{T}^{t_0}\mA\frac{\partial f}{\partial \theta}\frac{d\theta}{d\phi}dt.
\end{equation}
Using the conditions $\mA(T) = -\mI$, $\vh(t_0) = \vh_{t_0}$, $\vh(T) = \vh_T$, we 
rewrite the equation 
{\eqref{eq:A113}}
as
\begin{equation}\label{eq:A114}
\frac{d\vh_T}{d\phi} = - \mA(t_0)\frac{d\vh_{t_0}}{d\phi} + \int_{T}^{t_0}\mA\frac{\partial f}{\partial \theta}\frac{d\theta}{d\phi}dt.
\end{equation} 
Substituting 
{$\phi = \vh_{t_0}$ and $\phi = \theta$ respectively in \eqref{eq:A114} leads to} 
\begin{equation}
\frac{d\vh_T}{d\vh_{t_0}} = - \mA(t_0),
\quad
\frac{d\vh_T}{d\theta} = \int_{T}^{t_0}\mA\frac{\partial f}{\partial \theta}dt.
\end{equation} 
This proof is adapted from the proof provided by \cite{norcliffe2020_sonode} for general second-order neural ODEs by 
{differentiation and this proof only holds for $f\in C^1$.} 

\subsubsection{Proof of the First-Order Adjoint Sensitivity Equations: Integration Approach}\label{integraion-proof}
{The proof in 
Appendix~\ref{differentiation-proof} requires that $f\in C^1$. However,}
with activation functions like ReLU, $f$ may not be smooth enough to satisfy this requirement. 
Meanwhile, the adjoint equation 
{\eqref{eq:Appendix-NODE-Adjoint}}
that $\mA$ satisfies may not 
{have a continuous right hand side,} 
which can fail the Picard-Lindel\"{o}f theorem that 
guarantees the existence and uniqueness of solutions to the adjoint equation. 

{To circumvent these deficiencies, we propose a new proof based on integration.}
Assume that \(f(\vh, t, \theta)\) is continuous in \(t\) and
Lipschitz continuous in \(\vh, \theta\), and there exists
some open ball around \(\vh_{t_0} = \vs_0\), \(\theta=\theta_0\) such that for every pair of initial condition and parameters in the open ball, there exists a unique solution for $t\in [t_0,T]$. We denote the solution starting from \(\vh_{t_0} = \vs_0\), \(\theta=\theta_0\) as \(\vh_0\). In order to avoid difficulties in proving the existence and uniqueness of the solution, we explicitly define the adjoint equation through the following matrix exponential
\begin{equation}
\mA(t) = -\exp\Bigg\{-\int_{T}^t \frac{\partial f}{\partial\vh}(\vh_0(\tau), \tau, \theta_0)d\tau\Bigg\}.
\label{eq:int-A-def}
\end{equation} 
By definition, \(\mA\) is Lipschitz continuous and satisfies the 
differential equation almost everywhere
\begin{equation}\label{eq:Appendix-NODE-Adjoint-A.1.2}
\frac{d\mA(t)}{dt} = -\mA(t)\frac{\partial f}{\partial\vh}(\vh_0(t), t, \theta_0).
\end{equation} 
{Since \(\vh\in C^1(t)\) and \(\frac{d\mA}{d t}\in L^1(t)\), we obtain the following using integration by parts,}
\begin{equation}\label{eq:A123}
\mA\vh|_{t_0}^{T} = \int_{t_0}^{T} \mA\frac{\partial \vh}{\partial t}dt + \int_{t_0}^{T} \frac{d\mA}{dt}\vh dt.
\end{equation} 
Taking partial derivatives with respect to \(\phi\) on both sides 
{of \eqref{eq:A123}}, as \(\mA(t)\) is only a function of \(t\), we have
\begin{equation}\label{eq:136}
\mA\frac{\partial \vh}{\partial \phi}\Big|_{t_0}^{T} = \frac{\partial}{\partial \phi}\int_{t_0}^{T} \mA\frac{\partial \vh}{\partial t}dt + \frac{\partial}{\partial \phi}\int_{t_0}^{T} \frac{d\mA}{dt}\vh dt.
\end{equation} 
In order to exchange integral and derivatives, we use the dominated convergence theorem. Because \(f\) is Lipschitz continuous on $\vh$, \(\vh\) is Lipschitz continuous on \(\phi\), and thus \(\frac{\partial \vh}{\partial \phi}\) is Lebesgue integrable. Therefore, by chain rule, the following equation holds almost everywhere,
\begin{equation}\label{eq:137}
    \frac{\partial^2\vh}{\partial t\partial \phi} = \frac{\partial^2\vh}{\partial \phi\partial t} = \frac{d f}{d \phi} = \frac{\partial f}{\partial \theta}\frac{d \theta}{d \phi} + \frac{\partial f}{\partial \vh}\frac{d \vh}{d \phi}.
\end{equation}
Because $t$ is bounded, the right hand side of equation \eqref{eq:137} is Lebesgue integrable, and so is the left hand side. Because both \(\frac{\partial \vh}{\partial \phi}\) and \(\frac{\partial^2\vh}{\partial t\partial \phi}\) are Lebesgue integrable, by dominated convergence theorem, we have the following exchange of integrals and derivatives 
\begin{equation}\label{eq:138}
\frac{\partial}{\partial \phi} \int_{t_0}^{T} \mA\frac{\partial \vh}{\partial t} dt = \int_{t_0}^{T} \mA\frac{\partial^2 \vh}{\partial t \partial \phi} dt,
\quad 
\frac{\partial}{\partial \phi} \int_{t_0}^{T} \frac{d \mA}{d t}\vh dt = \int_{t_0}^{T} \frac{d \mA}{d t}\frac{\partial \vh}{\partial \phi} dt.
\end{equation} 
Combining equation \eqref{eq:136} with \eqref{eq:138} gives us 
\begin{equation}\label{eq:139}
\mA\frac{\partial \vh}{\partial \phi}\Big|_{t_0}^{T} = \int_{t_0}^{T} \mA\frac{\partial^2 \vh}{\partial t \partial \phi} dt + \int_{t_0}^{T} \frac{d \mA}{d t}\frac{\partial \vh}{\partial \phi} dt.
\end{equation} 

By taking Lebesgue integral of equation \eqref{eq:137}, we have the equation
\begin{equation}\label{eq:140}
\int_{t_0}^{T} \mA\frac{\partial^2 \vh}{\partial t \partial \phi} dt = \int_{t_0}^{T} \mA\Big(\frac{\partial f}{\partial \theta}\frac{d \theta}{d \phi} + \frac{\partial f}{\partial \vh}\frac{d \vh}{d \phi}\Big)dt.
\end{equation} 
{Meanwhile, at $\vh_0$, we can integrate equation \eqref{eq:Appendix-NODE-Adjoint-A.1.2} to a similar form as}
\begin{equation}\label{eq:141}
\int_{t_0}^{T} \frac{d \mA}{d t}\frac{\partial \vh}{\partial \phi} dt = -\int_{t_0}^{T} \mA \frac{\partial f}{\partial \vh}\frac{d \vh}{d \phi}dt.
\end{equation} 
Consequently, at \(\vh_0\), we can sum up equations \eqref{eq:139}, \eqref{eq:140}, and \eqref{eq:141} and arrive at
\begin{equation}
\mA\frac{\partial \vh}{\partial \phi}\Big|_{t_0}^{T} = \int_{t_0}^{T} \mA\frac{\partial f}{\partial \theta}\frac{d \theta}{d \phi}dt,
\end{equation} 
which is the same integral equation 
as 
{equation \eqref{eq:A113}} in the differentiation proof in Appendix \ref{differentiation-proof}. Thus, plugging in the initial conditions provides us with the same result.

\subsubsection{Corollary of the First-order Gradient Propagation}\label{1st-gradient-corollary}

An immediate corollary of the above proof is that combining equations \eqref{eq:1st-order-A} and \eqref{eq:int-A-def} results in
\begin{equation}\label{eq:cor-A.1.3}
    \frac{d\vh_T}{d\vh_{t_0}} = -\mA(t_0) = \exp\Bigg\{-\int_{T}^{t_0} \frac{\partial f}{\partial\vh}(\vh_0(\tau), \tau, \theta_0)d\tau\Bigg\}.
\end{equation}
As \eqref{eq:cor-A.1.3} is true for every choice of $t_0$, we can also generalize it to 
\begin{equation}\label{eq:144}
    \frac{d\vh_T}{d\vh_{t}} = \exp\Bigg\{-\int_{T}^{t} \frac{\partial f}{\partial\vh}(\vh_0(\tau), \tau, \theta_0)d\tau\Bigg\},
\end{equation}
which shows the relative gradient between different 
{times}
in integral.

\subsection{Second-order Adjoint Sensitivity Equation}

A SONODE satisfies the following equations
\begin{equation}
\frac{\partial \vh}{\partial t} = \vv,
\quad 
\frac{\partial \vv}{\partial t} = f(\vh(t), \vv(t), t, \theta),
\quad 
\vh(t_0) = \vh_{t_0},
\quad 
\vv(t_0) = \vv_{t_0},
\end{equation}
which can be viewed as a coupled first-order ODE 
{system} of the form

\begin{equation}
\frac{\partial }{\partial t}\begin{bmatrix}
\vh \\ \vv
\end{bmatrix} = \begin{bmatrix}
\vv \\ f(\vh(t), \vv(t), t, \theta)
\end{bmatrix},
\quad 
\begin{bmatrix}
\vh \\ \vv
\end{bmatrix}(t_0) = \begin{bmatrix}
\vh_{t_0} \\ \vv_{t_0}
\end{bmatrix}.
\end{equation}

Denote 
$\vz = \begin{bmatrix}\vh\\\vv\end{bmatrix}$ and final state as 

\begin{equation}
\begin{bmatrix}
\vh(T) \\ \vv(T)
\end{bmatrix}
= \begin{bmatrix}
\vh_{T} \\ \vv_{T}
\end{bmatrix} = \vz_T.
\end{equation}

Using the conclusions from Appendix \ref{1st-order-proofs}, then the adjoint equation 
{is given by}
\begin{equation}
\frac{\partial \mA(t)}{\partial t} = -\mA(t) \begin{bmatrix}
{\bf 0} & \mI \\
\frac{\partial f}{\partial \vh} & \frac{\partial f}{\partial \vv}
\end{bmatrix},
\quad 
\mA(T) = -\mI,
\quad 
\va(t) = -\frac{d\mathcal{L}}{d\vz_T} \mA(t).
\end{equation}

By rewriting $\mA = \begin{bmatrix}
\mA_\vh & \mA_\vv
\end{bmatrix}$, we have the following differential equations

\begin{equation}
\frac{\partial \mA_\vh(t)}{\partial t} = -\mA_\vv(t) \frac{\partial f}{\partial \vh},
\quad 
\frac{\partial \mA_\vv(t)}{\partial t} = -\mA_\vh(t) - \mA_\vv(t) \frac{\partial f}{\partial \vv},
\end{equation}

with initial conditions 

\begin{equation}
\mA_\vh(T) = -\begin{bmatrix}\mI\\{\bf 0}\end{bmatrix},
\quad 
\mA_\vv(T) = -\begin{bmatrix}{\bf 0}\\\mI\end{bmatrix},
\end{equation}

and adjoint states 

\begin{equation}
\va_\vh(t) = \frac{d\mathcal{L}}{d\vz_T} \mA_\vh(t),
\quad 
\va_\vv(t) = \frac{d\mathcal{L}}{d\vz_T} \mA_\vv(t).
\end{equation}

The gradient equations becomes 
\begin{equation}
\frac{d \mathcal{L}}{d \theta} = \int_{t_0}^{T}\va \begin{bmatrix}
{\bf 0} \\ \frac{\partial f}{\partial \theta} 
\end{bmatrix} dt = \int_{t_0}^{T}\va_\vv \frac{\partial f}{\partial \theta} dt,
\quad 
\frac{d \mathcal{L}}{d \vh_{t_0}} = \va_\vh (t_0),
\quad 
\frac{d \mathcal{L}}{d \vv_{t_0}} = \va_\vv (t_0).
\end{equation}

In SONODE, $\vh_{t_0}$ is fixed, and thus $\va_\vh$ disappears in gradient computation. Therefore, we are only interested in $\va_\vv$.
Thus the adjoint $\mA_\vv$ satisfies the 
following 
second-order ODE
\begin{equation}
\frac{\partial^2 \mA_\vv(t)}{\partial t^2} = \mA_\vv(t) \frac{\partial f}{\partial \vh} - \frac{\partial (\mA_\vv(t) \frac{\partial f}{\partial \vv})}{\partial t},
\end{equation}

and thus 

\begin{equation}\label{eq:254}
\frac{\partial^2 \va_\vv(t)}{\partial t^2} = \va_\vv(t) \frac{\partial f}{\partial \vh} - \frac{\partial (\va_\vv(t) \frac{\partial f}{\partial \vv})}{\partial t},
\end{equation}

with initial conditions

\begin{equation}\label{eq:255}
\va_\vv (T) = -\frac{d\mathcal{L}}{d\vz} \mA_\vv(T) = \frac{d\mathcal{L}}{d\vv_{T}}, 
\quad 
\frac{\partial \va_\vv(T)}{\partial t} = -\frac{d\mathcal{L}}{d\vh_{T}} - \va_\vv(T) \frac{\partial f}{\partial \vv}(T).
\end{equation}

This proves the second order adjoint equations for $\va_\vv$. 

\section{Proof of Propositions~\ref{prop:adjoint-HBNODE}, \ref{prop:adjoint-HBNODE-1st}, and \ref{prop:adjoint-GHBNODE}}

\subsection{Proof of Adjoint Equation for HBNODE (Propositions~\ref{prop:adjoint-HBNODE})}
\label{proof-adjoint-2nd-order-hbnode}

As HBNODE takes the form 

\begin{equation}
\frac{d^2\vh(t)}{dt^2} + \gamma \frac{d\vh(t)}{dt} = f(\vh(t),t,\theta), 
\end{equation}

which can also be viewed as a SONODE. By applying the adjoint equation of SONODE \eqref{eq:254}, 
we arrive at

\begin{equation}
\frac{\partial^2 \va(t)}{\partial t^2} = \va(t) \frac{\partial f}{\partial \vh} +\gamma \frac{\partial \va(t)}{\partial t}.
\end{equation}

As HBNODE only carries its state $\vh$ to the loss $\mathcal{L}$, we have $\frac{d\mathcal{L}}{d\vv_{T}} = 0$, and thus the initial conditions in equation \eqref{eq:255} 
become 

\begin{equation}
\va(T) = {\bf 0}, 
\quad 
\frac{\partial \va(T)}{\partial t} = -\frac{d\mathcal{L}}{d\vh_{T}} .
\end{equation}

\subsection{Proof of Adjoint Equation for First-order HBNODE (Proposition \ref{prop:adjoint-HBNODE-1st})}
\label{proof-adjoint-1st-order-hbnode}

The coupled form of HBNODE is a coupled first-order ODE 
{system} of the form

\begin{equation}
\frac{\partial }{\partial t}\begin{bmatrix}
\vh \\ \vm
\end{bmatrix} = \begin{bmatrix}
\vm \\ -\gamma \vm + f(\vh(t), t, \theta)
\end{bmatrix},
\quad 
\begin{bmatrix}
\vh \\ \vm
\end{bmatrix}(t_0) = \begin{bmatrix}
\vh_{t_0} \\ \vm_{t_0}
\end{bmatrix}.
\end{equation}

Denote the final state as 

\begin{equation}
\begin{bmatrix}
\vh(T) \\ \vm(T)
\end{bmatrix}
= \begin{bmatrix}
\vh_{T} \\ \vm_{T}
\end{bmatrix} = z.
\end{equation}

Using the conclusions from Appendix \ref{1st-order-proofs}, we have the adjoint equation 

\begin{equation}
\frac{\partial \mA(t)}{\partial t} = -\mA(t) \begin{bmatrix}
{\bf 0} & \mI \\
\frac{\partial f}{\partial \vh} & -\gamma \mI
\end{bmatrix},
\quad 
\mA(T) = -\mI,
\quad 
\va (t) = -\frac{d\mathcal{L}}{d\vz} \mA(t).
\end{equation}

Let $\begin{bmatrix}\va_\vh & \va_\vm\end{bmatrix} = \va$, by linearity we have 

\begin{equation}
\frac{\partial \begin{bmatrix}\va_\vh & \va_\vm\end{bmatrix}}{\partial t} = -\begin{bmatrix}\va_\vh & \va_\vm\end{bmatrix} \begin{bmatrix}
{\bf 0} & \mI \\
\frac{\partial f}{\partial \vh} & -\gamma \mI
\end{bmatrix},
\quad 
\begin{bmatrix}\va_\vh(T) & \va_\vm(T)\end{bmatrix} = \begin{bmatrix}\frac{d\mathcal{L}}{d\vh_{T}} & \frac{d\mathcal{L}}{d\vm_{T}}\end{bmatrix},
\end{equation}
which gives us the initial conditions at $t=T$,
and the simplified first-order ODE system 

\begin{equation}
\frac{\partial \va_\vh}{\partial t} = - \va_\vm\frac{\partial f}{\partial \vh},
\quad 
\frac{\partial \va_\vm}{\partial t} = -\va_\vh + \gamma \va_\vm.
\end{equation}

\subsection{Proof of Adjoint Equation for GHBNODE (Proposition~\ref{prop:adjoint-GHBNODE})}
\label{proof-adjoint-ghbnode}

The coupled form of GHBNODE is a 
first-order ODE system of the form

\begin{equation}\label{eq:264}
\frac{\partial }{\partial t}\begin{bmatrix}
\vh \\ \vm
\end{bmatrix} = \begin{bmatrix}
\sigma(\vm) \\ -\gamma \vm + f(\vh(t), t, \theta) - \xi \vh(t)
\end{bmatrix},
\quad 
\begin{bmatrix}
\vh \\ \vm
\end{bmatrix}(t_0) = \begin{bmatrix}
\vh_{t_0} \\ \vm_{t_0}
\end{bmatrix}.
\end{equation}

Denote the final state as 

\begin{equation}
\begin{bmatrix}
\vh(T) \\ \vm(T)
\end{bmatrix}
= \begin{bmatrix}
\vh_{T} \\ \vm_{T}
\end{bmatrix} = \vz_T.
\end{equation}

Using the conclusions from Appendix \ref{1st-order-proofs}, we have the adjoint equation 
\begin{equation}
\frac{\partial \mA(t)}{\partial t} = -\mA(t) \begin{bmatrix}
{\bf 0} & \sigma'(\vm) \\
\frac{\partial f}{\partial \vh}-\xi\mI & -\gamma \mI
\end{bmatrix},
\quad 
\mA(T) = -\mI,
\quad 
\va (t) = -\frac{d\mathcal{L}}{d\vz_T} \mA(t).
\end{equation}

Let $\begin{bmatrix}\va_\vh & \va_\vm\end{bmatrix} = \va$, by linearity we have 

\begin{equation}
\frac{\partial \begin{bmatrix}\va_\vh & \va_\vm\end{bmatrix}}{\partial t} = -\begin{bmatrix}\va_\vh & \va_\vm\end{bmatrix} \begin{bmatrix}
{\bf 0} & \sigma'(\vm) \\
\frac{\partial f}{\partial \vh}-\xi\mI & -\gamma \mI
\end{bmatrix},
\quad 
\begin{bmatrix}\va_\vh(T) & \va_\vm(T)\end{bmatrix} = \begin{bmatrix}\frac{d\mathcal{L}}{d\vh_{T}} & \frac{d\mathcal{L}}{d\vm_{T}}\end{bmatrix},
\end{equation}
which gives us the initial conditions at $t=T$,
and the simplified first-order ODE system
\begin{equation}
\frac{\partial \va_\vh}{\partial t} = - \va_\vm \Big(\frac{\partial f}{\partial \vh}-\xi\mI\Big),
\quad 
\frac{\partial \va_\vm}{\partial t} = -\va_\vh \sigma'(\vm) + \gamma \va_\vm.
\end{equation}

\section{Vanishing and Exploding Gradients in Training RNNs}
\label{appendix-vanishing-gradent}
Recurrent cells are the building blocks of RNNs. A recurrent cell can be mathematically written as
\begin{equation}\label{eq:RNN:Cell}
\vh_t = \sigma(\mU\vh_{t-1} + \mW\vx_t + \vb),\ \vx_t \in \RR^d,\ 
\mbox{for}\ t=1, 2, \cdots, T,
\end{equation}
where $\vh_t\in \RR^h$ is the hidden state, $\mU \in \RR^{h\times h}, \mW \in \RR^{h\times d}$, and $\vb\in \RR^h$ are trainable parameters; $\sigma(\cdot)$ is a nonlinear activation function, e.g., sigmoid. 
Backpropagation through time is a popular algorithm for training RNNs, which usually results in exploding or vanishing gradients \citep{bengio1994learning}. Thus RNNs may fail to learn 
long term dependencies. As an illustration, let $\vh_T$ and $\vh_t$ be the state vectors at the timestamps $T$ and $t$ ($T\gg t$), respectively. Assume $\mathcal{L}$ is the loss to minimize, then 
\begin{equation}
\label{eq:gradient:rnn}
{\small \frac{\partial \mathcal L}{\partial \vh_t} = \frac{\partial \mathcal{L}}{\partial\vh_T}\cdot\frac{\partial \vh_T}{\partial \vh_t} = \frac{\partial \mathcal L}{\partial \vh_T}\cdot \prod_{k=t}^{T-1}\frac{\partial \vh_{k+1}}{\partial \vh_k} = \frac{\partial \mathcal L}{\partial \vh_T}\cdot \prod_{k=t}^{T-1}(\mD_k\mU^\top),}
\end{equation}
where 
{\small $\mD_k = {\rm diag}(\sigma'(\mU\vh_k + \mW\vx_{k+1}))$} is a diagonal matrix with $\sigma'(\mU\vh_k + \mW\vx_{k+1})$ being its diagonal entries. 
{\small $\|\prod_{k=t}^{T-1}(\mD_k\mU^\top)\|_2$} tends to either vanish or explode \cite{bengio1994learning}.

When applying RNNs to sequence applications with 
$\vx = (\vx_1, \cdots, \vx_T)$ 
be 
an input sequence of length $T$ and $\vy = (y_1, \cdots, y_T)$ be the sequence of labels,
we let $\mathcal{L}_t$ be the loss at the timestamp $t$ and the total loss on the whole sequence be
\begin{equation}
\label{eq:whole:loss}
\mathcal{L} = \sum_{t=1}^T \mathcal{L}_t,
\end{equation}
the vanishing or exploding issue can be shown 
{following}
\eqref{eq:gradient:rnn}.

For neural ODEs, note that the adjoint state $\va(t)$ is defined as $\partial\mathcal{L}/\partial \vh(t)$, which also tends to explode or vanish during training. 

\subsection{Derivation of equation \eqref{eq:HBNODE-gradient}}

GHBNODE can be viewed as a system of higher dimensional NODE as in equation \eqref{eq:264}. With $\vz = \begin{bmatrix}\vh \\ \vm\end{bmatrix}$, $\vz$ satisfies NODE equation, and therefore it also satisfies the equation for relative gradient information as in \eqref{eq:144},
\begin{equation}\label{eq:272}
    \frac{d\vz_T}{d\vz_{t}} = \exp\Bigg\{-\int_{T}^{t} \frac{\partial f}{\partial\vz}(\vz_0(\tau), \tau, \theta_0)d\tau\Bigg\}. 
\end{equation}
By definition of multivariate derivatives, we have
\begin{equation}\label{eq:273}
    \frac{d\vz_T}{d\vz_{t}} = \begin{bmatrix}
\frac{\partial\vh_T}{\partial\vh_t} & \frac{\partial\vh_T}{\partial\vm_t}\\
\frac{\partial\vm_T}{\partial\vh_t} &\frac{\partial\vm_T}{\partial\vm_t}\\
\end{bmatrix}, 
\end{equation}
and 
\begin{equation}\label{eq:274}
\frac{\partial f}{\partial\vz} =
\begin{bmatrix}
{\bf 0} & \frac{\partial \sigma}{\partial \vm}\\
\frac{\partial f}{\partial\vh}-\xi\mI &-\gamma\mI
\end{bmatrix}.
\end{equation}
With equations \eqref{eq:273} and \eqref{eq:274}, we can rewrite equation \eqref{eq:272} in terms of $\vh$ and $\vm$ as 
\begin{equation}
\begin{bmatrix}
\frac{\partial\vh_T}{\partial\vh_t} & \frac{\partial\vh_T}{\partial\vm_t}\\
\frac{\partial\vm_T}{\partial\vh_t} &\frac{\partial\vm_T}{\partial\vm_t}\\
\end{bmatrix}=
\exp\left\{-\int_T^t\begin{bmatrix}
{\bf 0} & \frac{\partial \sigma}{\partial \vm}\\
\frac{\partial f}{\partial\vh}-\xi\mI &-\gamma\mI
\end{bmatrix} \right\}ds.
\end{equation}
In particular, since HBNODEs are GHBNODEs with $\xi=0$ and $\sigma$ being the identity map, the gradient equation of HBNODEs takes the form 
\begin{equation}
\begin{bmatrix}
\frac{\partial\vh_T}{\partial\vh_t} & \frac{\partial\vh_T}{\partial\vm_t}\\
\frac{\partial\vm_T}{\partial\vh_t} &\frac{\partial\vm_T}{\partial\vm_t}\\
\end{bmatrix}=
\exp\left\{-\int_T^t\begin{bmatrix}
{\bf 0} & \mI\\
\frac{\partial f}{\partial\vh} &-\gamma\mI
\end{bmatrix} \right\}ds.
\end{equation}
{This concludes the derivation of equation \eqref{eq:HBNODE-gradient}.}

\section{Proof of Proposition~\ref{lemma-eigan-M}}

\begin{proof}
Let ${\mF} = \frac{1}{t-T}\int_T^t\frac{\partial f}{\partial \vh}(\vh(s), s, \theta)ds - \xi\mI$, ${\mJ} = \frac{1}{t-T}\int_T^t\frac{\partial \sigma}{\partial \vm}(\vm(s))ds$, and ${\mH} = \frac{1}{t-T} \mM$, then we have the following equation
\begin{equation}
    {\mH} = \frac{1}{t-T}\mM = \begin{bmatrix}0 & {\mJ} \\ {\mF} & -\gamma \mI\end{bmatrix}.
\end{equation}
As $(\lambda+\gamma)\mI$ commutes with any matrix ${\mF}$, the characteristics polynomials of ${\mH}$ and ${\mJ\mF}$ 
{satisfy} the
relation
\begin{equation}
    ch_{\mH}(\lambda) = \det(\lambda \mI - {\mH}) = \det \begin{bmatrix}\lambda \mI & -{\mJ} \\ -{\mF} & (\lambda + \gamma) \mI\end{bmatrix} = det(\lambda(\lambda + \gamma)\mI - {\mJ\mF}) = -ch_{\mJ\mF}(\lambda(\lambda+\gamma)).
\end{equation}
Since the characteristics polynomial of ${\mJ\mF}$ splits in the field $\mathbb{C}$ of complex numbers, i.e. $ch_{\mJ\mF}(x) = \prod_{i=1}^n (x - \lambda_{{\mJ\mF},i})$, we have
\begin{equation}
    ch_{\mH}(\lambda) = -ch_{\mJ\mF}(\lambda(\lambda+\gamma)) = -\prod_{i=1}^n (\lambda(\lambda+\gamma) - \lambda_{{\mJ\mF},i}).
\end{equation}
Therefore, the eigenvalues of ${\mH}$ 
{appear} in $n$ pairs with each pair 
{satisfying}
the 
quadratic equation 
\begin{equation}
    \lambda(\lambda+\gamma) - \lambda_{{\mJ\mF},i} = 0.
\end{equation}
By Vieta's formulas, 
the sum of these pairs are all $-\gamma$. Therefore, the eigenvalues of $\mM$ comes in $n$ pairs and the sum of each pair is 
$-(t-T)\gamma$.
\end{proof}

\section{Experimental details 
}\label{appendix:experimental-details}

We first list some common settings below:
\begin{itemize}
    \item NODE and ANODE do not have initial layers.
    \item For SONODE $n^* = 2n$, and for other ones $n^*=n$.
    \item Hyper parameters are listed in Table \ref{Tab:hyperparameters}
    \item HTanh: HardTanh(-5, 5)
    \item LReLU: LeakyReLU(0.3)
    \item tpad: Padding with time $t$ within ODE. i.e., transform the shape $c\times x\times y$ to $(c+1)\times x\times y$ by concatenating with a tensor of shape $1\times x\times y$ filled with all $t$.
    \item For all tasks, we use learnable $\gamma$ with $\epsilon=1$ for both HBNODE and GHBNODE, and learnable $\xi$.
    \item ${\rm fc}_n$: a fully connected layer with output dimension to be $n$.
\end{itemize}

\begin{table}[!ht]
\fontsize{8.0}{8.0}\selectfont
\centering
\begin{threeparttable}
\caption{The hyper-parameters for each models.}
\label{Tab:hyperparameters}
\begin{tabular}{cccccc}
\toprule[1.0pt]
\ \ \ Model\ \ \   &\ \ \ NODE\ \ \  &\ \ \  ANODE\ \ \   &\ \ \  SONODE\ \ \  & \ \ \  HBNODE\ \ \  & \ \ \  GHBNODE\ \ \  \cr
\midrule[0.8pt]
$n$ (Initialization)  & 1 & 2 & 1 & 1 & 1 \cr
$h$ (Initialization)  & 22 & 22 & 22 & 22 & 22 \cr
$n$ (Point Cloud)  & 2 & 3 & 2 & 2 & 2 \cr
$h$ (Point Cloud)  & 20 & 20 & 13 & 14 & 14 \cr
$n$ (MNIST)  & 1 & 6 & 5 & 5 & 6 \cr
$h$ (MNIST)  & 92 & 64 & 50 & 50 & 45 \cr
$n$ (CIFAR)  & 3 & 13 & 12 & 12 & 12 \cr
$h$ (CIFAR)  & 125 & 64 & 50 & 51 & 51 \cr
\bottomrule[1.0pt]
\end{tabular}
\end{threeparttable}
\end{table}

\begin{table}[!ht]
\fontsize{8.0}{8.0}\selectfont
\centering
\begin{threeparttable}
\caption{The hyper-parameters for ODE-RNN integration models.}
\label{Tab:hyperparameters2}
\begin{tabular}{cccccc}
\toprule[1.0pt]
Model  &ODE-RNN  & ANODE-RNN   & SONODE-RNN  &  HBNODE-RNN  &  GHBNODE-RNN  \cr
\midrule[0.8pt]
$d$ & 1 & 1 & 2 & 2 & 2 \cr
$n$ (Plane Vibration)  & 21 & 27 & 19 & 20 & 20 \cr
$h_1$ (Plane Vibration)  & 63 & 83 & 19 & 20 & 20 \cr
$h_2$ (Plane Vibration)  & 84 & 108 & 19 & 20 & 20 \cr
$n$ (Walker 2D)  & 24 & 24 & 23 & 24 & 24 \cr
$h_1$ (Walker 2D)  & 72 & 72 & 46 & 48 & 48 \cr
$h_2$ (Walker 2D)  & 48 & 48 & 46 & 48 & 48 \cr
\bottomrule[1.0pt]
\end{tabular}
\end{threeparttable}
\end{table}

\subsection{Network architecture used in Section~\ref{sec:GHBNODEs} Initialization Test}

\begin{itemize}
\item 
$
\text{ODE}: 
\text{input}_{n^*+1} 
\to \text{fc}_{n}
$
\end{itemize}

\subsection{Experimental details for \ref{sec:point-cloud}}
\begin{itemize}
\item 
$
\text{Initial Velocity}: 
\text{input}_{2} 
\to \text{fc}_{h}
\to \text{HTanh}
\to \text{fc}_{h}
\to \text{HTanh}
\to \text{fc}_{n}
$

\item 
$
\text{ODE}: 
\text{input}_{n^*} 
\to \text{fc}_{h}
\to \text{ELU}
\to \text{fc}_{h}
\to \text{ELU}
\to \text{fc}_{n}
$
\item 
$
\text{Output}: 
\text{input}_{n} 
\to \text{fc}_{1}
\to \text{Tanh}
$
\end{itemize}

\subsection{Experimental details for \ref{sec:image-classification}}

\subsubsection{MNIST}

\begin{itemize}
\item 
$
\text{Initial Velocity}: 
\text{input}_{1\times 28\times 28} 
\to \text{conv}_{h, 1}
\to \text{LReLU}
\to \text{conv}_{h, 3}
\to \text{LReLU}
\to \text{conv}_{2n-1, 1}
$
\item 
$
\text{ODE}: 
\text{input}_{n^*\times 28\times 28} 
\to \text{tpad}
\to \text{conv}_{h, 1}
\to \text{ReLU}
\to \text{tpad}
\to \text{conv}_{h, 3}
\to \text{ReLU}
\to \text{tpad}
\to \text{conv}_{n, 1}
$
\item 
$
\text{Output}: 
\text{input}_{n\times 28\times 28} 
\to \text{fc}_{10}
$
\end{itemize}

\subsubsection{CIFAR}

\begin{itemize}
\item 
$
\text{Initial Velocity}: 
\text{input}_{3\times 28\times 28} 
\to \text{conv}_{h, 1}
\to \text{LReLU}
\to \text{conv}_{h, 3}
\to \text{LReLU}
\to \text{conv}_{2n-3, 1}
$
\item 
$
\text{ODE}: 
\text{input}_{n^*\times 32\times 32} 
\to \text{tpad}
\to \text{conv}_{h, 1}
\to \text{ReLU}
\to \text{tpad}
\to \text{conv}_{h, 3}
\to \text{ReLU}
\to \text{tpad}
\to \text{conv}_{n, 1}
$
\item 
$
\text{Output}: 
\text{input}_{n\times 32\times 32} 
\to \text{fc}_{10}
$
\end{itemize}
\subsection{Experimental details for \ref{sec:physical-systems}}

\begin{itemize}
\item 
$
\text{ODE}: 
\text{input}_{n*} 
\to \text{fc}_{h_1}
\to \text{ReLU}
\to \text{fc}_{h_2}
\to \text{ReLU}
\to \text{fc}_{n}
$
\item 
$
\text{RNN}: 
\text{input}_{dn+k} 
\to \text{fc}_{dn}
$
\item 
$
\text{Output}: 
\text{input}_{n} 
\to \text{fc}_{5}
$
\end{itemize}

\subsection{Experimental details for \ref{sec:sequential-modeling}}

\begin{itemize}
\item 
$
\text{ODE}: 
\text{input}_{n*} 
\to \text{fc}_{n}
$
\item 
$
\text{RNN}: 
\text{input}_{dn+k} 
\to \text{fc}_{h_1}
\to \text{Tanh}
\to \text{fc}_{h_2}
\to \text{Tanh}
\to \text{fc}_{dn}
$
\item 
$
\text{Output}: 
\text{input}_{n} 
\to \text{fc}_{17}
$
\end{itemize}



\subsection{Experimental details for ODE vs. HBNODE on benchmarks}\label{appendix-compare-dynamics-odes}
To numerically show that the HBNODE \eqref{eq:Appendix-HBNODE2} converges faster to the stationary point than the ODE limit of gradient descent \eqref{eq:ODE-HBODE}, we apply the Dormand–Prince-45 ODE solver, which is the default solver for NODEs, to solve both ODEs. We set $F(\vx)$ to be two celebrated functions used in the optimization community, namely, the Rosenbrock and the Beale functions.


\paragraph{Rosenbrock function.} The Rosenbrock function is given by
$$
F(\vx):=F(x,y)=100(y-x^2)^2 + (1-x)^2,
$$
which has the minimum $(x,y)=(1,1)$. Starting from $(0,0)$, we apply Dormand–Prince-45 solver using a step size $\Delta t=0.001$ to solve 
both ODEs \eqref{eq:Appendix-HBNODE2} and \eqref{eq:ODE-HBODE} for $t$ from $0$ to $1$. For the HBODE, we set 
$\gamma=0.9$ and set the initial value of $d\vx/dt=(0,0)$. 


\paragraph{Beale function.} The Beale function is given by
$$
F(x,y)=(1.5-x+xy)^2 + (2.25-x+xy^2)^2 + (2.625-x+xy^3)^2 
$$
which has the minimum $(x,y)=(3,0.5)$. Starting from $(0,0)$, we apply Dormand–Prince-45 solver using a step size $0.01$ to solve 
both ODEs in \eqref{eq:Appendix-HBNODE2} and \eqref{eq:ODE-HBODE} for $t$ from $0$ to $2$. For the HBODE, we set 
$\gamma = 0.7$ and set the initial value of $d\vx/dt=(0,0)$.

\end{document}